\documentclass[11pt,a4paper]{article}
\usepackage{times,latexsym}
\usepackage{url}
\usepackage[T1]{fontenc}
\usepackage{tabularx}
\usepackage{booktabs}
\usepackage{amssymb}%
\usepackage{pifont}%
\newcommand{\cmark}{\ding{51}}%
\newcommand{\xmark}{\ding{55}}%

\usepackage[acceptedWithA]{tacl2021v1}

\usepackage{xspace,mfirstuc,tabulary}

\newif\iftaclinstructions
\taclinstructionsfalse %
\iftaclinstructions
\renewcommand{\confidential}{}
\renewcommand{\anonsubtext}{(No author info supplied here, for consistency with
TACL-submission anonymization requirements)}
\newcommand{\instr}
\fi

\iftaclpubformat %

\else

\fi

\usepackage{comment}
\usepackage{graphicx}
\usepackage{multirow}
\newcommand{\todo}[1]{\textcolor{black}{#1}}
\newcommand{\gog}[1]{\textcolor{black}{#1}}
\newcommand{\hw}[1]{\textcolor{black}{#1}}
\newcommand{\al}[1]{\textcolor{black}{#1}}
\newcommand{\alfinal}[1]{\textcolor{black}{#1}}
\newcommand{\hwfinal}[1]{\textcolor{black}{#1}}
\newcommand{\specialcell}[2][c]{%
  \begin{tabular}[#1]{@{}c@{}}#2\end{tabular}}

\usepackage{microtype}

\DeclareGraphicsExtensions{.pdf,.ai,.jpg,.png}
\setkeys{Gin}{pagebox=artbox}
\graphicspath{{./img/}}

\newcommand{\bsfigure}[3][]{%
	\begin{figure}[t]
		\centering
		\includegraphics[#1]{#2}
		\caption{#3}\label{#2}%
 	 \end{figure}
}

\newcommand{\hwfigure}[3][t!]{%
	\begin{figure*}[#1]
		\centering
		\includegraphics[scale=1.0]{#2}
    		\caption{#3}\label{#2}%
  	\end{figure*}
}

\RequirePackage{type1cm}%
\RequirePackage{color}
\RequirePackage{soul}
\setstcolor{blue}
\definecolor{violet}{rgb}{0.5,0.0,0.5}

\newsavebox\bscombox
\newcommand{\bscom}[3][]{%
  \sbox{\bscombox}{\fontsize{8}{9}\selectfont#1#2#3}
  \noindent
  \st{#2}{\selectfont
    \color{blue}#3\ifx\\#1\\\else{\fontsize{8}{9}\selectfont\color{violet}[#1]}\fi
    }
  }

\title{What kind of Knowledge is needed in Computational Argumentation?}
\title{Scientia Potentia Est: \\ On the State and the Fate of Knowledge in Computational Argumentation}
\title{Scientia Potentia Est -- \\ On the Role of Knowledge in Computational Argumentation}

\author{
 Anne Lauscher,\Thanks{~Equal contribution.} \textsuperscript{1} Henning Wachsmuth,\footnotemark[1] \textsuperscript{2} Iryna Gurevych,\textsuperscript{3} and Goran Glava\v{s}\textsuperscript{4} \\
 \textsuperscript{1}MilaNLP, Bocconi University, Italy\\
 \textsuperscript{2}Department of Computer Science, Paderborn University, Germany \\
 \textsuperscript{3}Ubiquitous Knowledge Processing Lab, TU Darmstadt, Germany \\
\textsuperscript{4}CAIDAS, University of Würzburg, Germany\\
  \small{\tt anne.lauscher@unibocconi.it,\,henningw@upb.de},\\ \small{\tt gurevych@ukp.informatik.tu-darmstadt.de,\,goran.glavas@uni-wuerzburg.de} \\
}

\date{}

\begin{document}
\maketitle
\begin{abstract}
Despite extensive research efforts in recent years, computational argumentation (CA) remains one of the most challenging areas of natural language processing. The reason for this is the inherent complexity of the cognitive processes behind human argumentation, which integrate a plethora of different types of knowledge, ranging from topic-specific facts and common sense to rhetorical knowledge. The integration of knowledge from such a wide range in CA requires modeling capabilities far beyond many other natural language understanding tasks. Existing research on mining, assessing, reasoning over, and generating arguments largely acknowledges that much more knowledge is needed to accurately model argumentation computationally. However, a systematic overview of the types of knowledge introduced in existing CA models is missing, hindering targeted progress in the field. Adopting the operational definition of knowledge as any task-relevant normative information not provided as input, the survey paper at hand fills this gap by (1) proposing a taxonomy of types of knowledge required in CA tasks, (2) systematizing the large body of CA work according to the reliance on and exploitation of these knowledge types for the four main research areas in CA, and (3) outlining and discussing directions for future research efforts in CA.
\end{abstract}

\section{Introduction}
The phenomenon of argumentation, a direct reflection of human reasoning in natural language, has fascinated scholars across societies and cultures since the ancient times~\citep{kennedy2007aristotle,lloyd:2007}. The computational modeling of human argumentation, commonly referred to as \emph{computational argumentation~(CA)}, has evolved into one of the most prominent and at the same time most challenging areas in natural language processing (NLP)~\citep{lippi2015argument}.\,

CA encompasses several families of tasks and research directions, the main ones in NLP being argument \textit{mining}, \textit{assessment}, \textit{reasoning}, and \textit{generation}. Although it bears some resemblance to other NLP tasks, such as opinion mining and natural language inference (NLI), it is widely acknowledged to be of much higher difficulty than the other tasks~\citep{habernal2014argumentation}. While opinion mining \cite{liu2012sentiment} assesses stances towards entities or controversies by asking \emph{what} the opinions are, CA provides answers to a more difficult question: \emph{why} is the stance of an opinion holder the way it is? In a similar vein, while NLI focuses on detecting simple entailments between statement pairs \cite{bowman2015large,dagan2013recognizing}, CA addresses more complex reasoning scenarios that involve multiple entailment steps, often over implicit premises \cite{boltuzic-snajder-2016-fill}. 

CA targets reasoning processes that are only partially explicated in text. Its mastery thus requires advanced natural language understanding capabilities and a substantial amount of background knowledge \citep{moens2018argumentation,paul2020argumentative}. For example, the assessment of an argument's quality not only depends on the actual content of an argumentative text or speech but also on social and cultural context, such as speaker and audience characteristics, including their individual values, ideologies, and relationships~\citep{wachsmuth-etal-2017-computational}. Such contextual information remains most often implicit. For any concrete CA task, we here refer to all information that is not explicitly provided as input to models tackling the task but is (potentially) useful for it and (in most cases) normative in nature as \emph{knowledge} (we detail this notion in \S\ref{sec:knowledge}).

Although there is ample awareness of the need for integrating various types of knowledge in CA models in the research community, there is no systematic overview of the types of knowledge that existing models and solutions for the different CA tasks rely on. This impedes targeted progress in pressing subareas of CA, such as argument generation. While general surveys on CA~\citep[e.g.,][]{cabrio2018five, lawrence2020argument} and its subareas~\citep[e.g.,][]{al-khatib-etal-2021-argument, schaefer2021argument} represent good starting points for targeted research along these lines, they lack a systematic analysis of the roles that different types of knowledge play in different CA tasks.

\paragraph{Contributions.} 

In this work, we aim to systematically inform the research community  about the types of knowledge that have---or have not yet---been integrated into computational models in different CA tasks. For this purpose, we \textbf{(1)}~propose a pyramid-like taxonomy systematizing the relevant types of knowledge. The pyramid is organized by knowledge specificity, from linguistic knowledge and world and topic knowledge to argumentation-specific and task-specific knowledge. Starting from 162 CA publications, we \textbf{(2)} survey the existing body of work with respect to the level of integration of the various types of knowledge and respective methodology by which the knowledge of each type is integrated into models. To this end, we carry out an expert annotation study in which we manually label individual papers with the types from the knowledge pyramid. Finally, we \textbf{(3)}~identify trends and challenges in the four most prominent CA subareas (mining, assessment, reasoning, and generation), summarizing them into three key recommendations for future CA research:
\begin{enumerate}
\setlength{\itemsep}{0pt}
\item
{\em All CA tasks are expected to benefit from more modeling of world and topic knowledge.} Although several studies report empirical gains from incorporating these types of knowledge, their inclusion is still an exception rather than a rule across the landscape of all CA tasks. 
\item
{\em Argument mining tasks are expexted to benefit from more modeling of argumentation- and task-specific knowledge.} Such specialized knowledge has been proven effective in assessment, reasoning, and generation tasks. Yet, it has so far been exploited only sporadically in argument mining approaches.
\item
{\em All CA tasks are expected to benefit from applying key techniques to other types of knowledge and data.} As an example, methods that represent symbolic input in a semantic vector space (e.g., pretrained word embeddings or language models) are still rarely applied to sources other than text (e.g., to knowledge bases). The bottleneck to a wider application of general-purpose techniques such as representation learning in CA is the lack of structured knowledge resources. We thus argue that significant progress in CA critically hinges on the availability of such resources at larger scale. Accordingly, based on the results of this survey effort, we strongly encourage the CA community to foster the creation of knowledge-rich argumentative corpora.     
\end{enumerate}

\paragraph{Structure.} 

We start with an overview of the field of CA and its four most prominent subareas (\S\ref{sec:background}). In \S\ref{sec:methodology}, we describe our survey methodology, before we establish the knowledge pyramid and present the results of the survey with respect to the types of knowledge from the pyramid (\S\ref{sec:taxonomy}). On this basis, we summarize emerging trends (\S\ref{sec:trends}) and offer recommendations for future progress in CA~(\S\ref{sec:future}).

\section{Background}
\label{sec:background}
The study of argumentation in Western societies can be traced back to Ancient Greece. With the development of democracy and, thereby, the need to influence public decisions, the art of convincing others became an essential skill for successful participation in the democratic process~\citep{kennedy2007aristotle}. In that period, rhetorical theories also started appearing in Eastern societies and cultures, such as {\em Nyaya Sutra}~\citep{lloyd:2007}. Since then, a plethora of phenomena in the realm of argumentation, such as fallacies \cite{hamblin:1970} and argumentation schemes~\citep{Reed2003-REEASI-5}, have been studied extensively, usually focusing on specific domains, such as science~\citep{nigel1977referencing} and law~\citep{toulmin_uses_2003}. 

With the growing amount of argumentative data available publicly in web debates, scientific articles, and other internet sources, the computational modeling of argumentation, \emph{computational argumentation (CA)}, gradually gained prominence and popularity in the NLP community.  As depicted in Figure~\ref{ca-overview}, CA can be divided into four main subareas that represent the main high-level types of tasks being tackled with computational models: mining, assessment, reasoning, and generation.

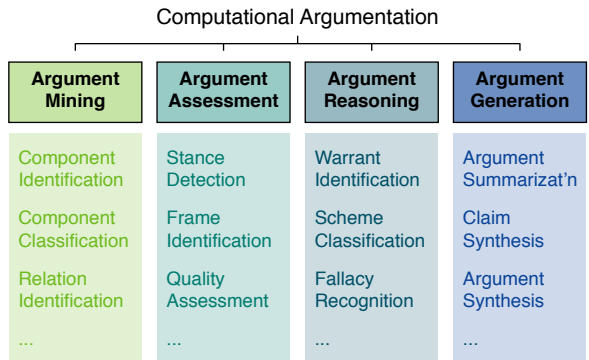
\bsfigure{ca-overview}{The four main subareas of computational argumentation (argument mining, argument assessment, argument reasoning, and argument generation) with three of their most prominent respective tasks each.}

\paragraph{Argument Mining.} 

Argument mining deals with the extraction of argumentative structures from natural language text~\citep[e.g.,][]{stab2017parsing}. Traditionally, it has been addressed with a pipeline of models each tackling one analysis task, most commonly \textit{component identification}, \textit{component classification}, and \textit{relation identification}~\citep{lippi2015argument}. The set of argument components and relations is defined by the selected underlying argument model which reflects the rhetorical, dialogical, or monological structure of argumentation~\citep{bentaharTaxonomyArgumentationModels2010}. 

For instance, the model of \newcite{toulmin_uses_2003}, designed for the legal domain, encompasses six  components: a claim with an optional qualifier, data (i.e., a fact supporting the claim) connected to the claim via a warrant (i.e., the reason why support is given) and its backing, and a rebuttal (i.e., a counterconsideration to the claim). Relations model the support or attack of components (or arguments) by others, sometimes with more fine-grained subtypes \cite{freeman:2011}. In contrast to argument reasoning (see below), the information needed for inferring argumentative relations is contained in the text.

\paragraph{Argument Assessment.} 

Computational models that address tasks in this subarea typically focus on particular properties of arguments in their context and automatically assign discrete or numeric labels for these properties. This includes the \emph{classification of stance} towards some target~\cite{bar-haim-etal-2017-stance} as well as the \emph{identification of frames} (or aspects) covered by the argument~\cite{ajjour-etal-2019-modeling}. Arguably, the most popular family of tasks belongs to argument \textit{quality assessment}, which has been studied under various conceptualizations, such as clarity~\citep{persing-ng-2013-clarity} or convincingness~\citep{Habernal2016WhichAI}. \citet{wachsmuth-etal-2017-computational} propose a taxonomy that divides the overall quality of an argument into three complementary aspects: logic, rhetoric, and dialectic. Each of these three aspects further consists of several quality dimensions (e.g., the dimension of global acceptability for the dialectical aspect).

\paragraph{Argument Reasoning.}  

In this subarea, the task is to understand the reasoning process behind an argument. In NLP, reasoning is instantiated in tasks such as predicting the entailment relationship between a premise and a hypothesis by means of natural language inference \citep{WilliamsNB17:mnli}, or the more complex task of \emph{warrant identification}, that is, to find (or even reconstruct) the missing warrant \citep{tian-etal-2018-ecnu}. Others have tried to \emph{classify schemes} of inferences happening in arguments \cite{feng-hirst-2011-classifying} or to \emph{recognize fallacies} of certain reasoning types in arguments, such as the common ad-hominem fallacy \cite{habernal-etal-2018-name,delobelle-etal-2019-computational}. 

In argument reasoning, the challenge lies in inducing additional knowledge---not explicated in the text---from existing components, as opposed to relation identification, which focuses on recognizing argumentative content present in the text. In other words, argument mining structures explicated arguments and their connections, whereas argument reasoning infers knowledge missing from the text (e.g., a warrant that connects the premise to the claim). \al{In practice, however, there is no guarantee that annotators for argument mining tasks (e.g., relation identification) do not resort to out-of-text reasoning, leveraging their commonsense and world knowledge to perform the task. However, from a structural point of view, a premise may still be given by an author to support a claim (e.g., indicated by lexical cues like \emph{because}), while from a reasoning perspective, the premise might be irrelevant to the claim (e.g., the claim does not logically follow from the given premise.}

\paragraph{Argument Generation.} 

With conversational AI (i.e., dialogue systems) arguably becoming the most prominent application in modern NLP and AI, the research efforts on generating argumentative language have also been gaining traction. Main tasks in argument generation include the \emph{summarization of arguments} given~\cite{wang-ling-2016-neural}, the \emph{synthesis of new claims} and other argument components~\cite{bilu-slonim-2016-claim}, and the \emph{synthesis of entire arguments}, possibly conforming to some rhetorical strategy~\citep{el-baff-etal-2019-computational}. 

The impact of argument generation is, for example, demonstrated by Project Debater~\cite{slonim2018project}, a well-known argumentation system which combines models for several generation tasks.

\section{Methodology}
\label{sec:methodology}
In this section, we first provide the definition of knowledge upon which we base this work. Then, we detail the methodology that we devised and pursued in order to organize the types of knowledge that CA approaches and models utilize.

\subsection{An Operational Definition of Knowledge}
\label{sec:knowledge}

Various definitions of ``knowledge'' have been proposed in the literature. One of the oldest is the tripartite definition of \citet{plato2019theaetetus}, who accepted as knowledge any \emph{justified true belief}. This definition was later often challenged as being too narrow and was, accordingly, extended~\citep[e.g.,][]{10.2307/2024268,10.2307/3070991}. As part of this effort, \citet{Dretske1981-DREKAT} dressed Plato's view into an information-theoretic gown, defining knowledge as \emph{information-caused belief}, \gog{specifying more narrowly the informational source of the belief as the only valid justification and de facto eliminating the veracity constraint}.   

Departing from attempts to define knowledge ontologically, \citet{gottschalk2013internet} \gog{adopt an impact-based viewpoint} and argue that it is more important to understand \emph{what knowledge can do} and \emph{what it is like} than to ontologically answer \textit{what knowledge is}. In their view, knowledge is thus normative and has practical implications. 
In the work at hand, we adopt this impact-oriented view on knowledge. We further operationalize the view, in the context of NLP and CA, as follows:

\medskip
\noindent 
{\textit{\textbf{Knowledge} is any kind of normative information that is considered to be relevant for solving a task at hand and that is not given as task input itself.}}

\medskip
\noindent 
In CA research, knowledge has been be modeled in a variety of forms that conform to this definition, ranging from lexicons, and engineered features to \gog{specially tailored pipelines, model components, or overall algorithm design (e.g., auxiliary tasks, or special training objectives)}. \gog{While this is not the primary dimension of our analysis (see \S\ref{sec:pyramid}), it is worth noting the difference between knowledge that is presented \textit{explicitly}, i.e., that can be rather directly used to shape the input representations for the task (e.g., lexicons, feature engineering, predictions of existing auxiliary models), and knowledge that is introduced \textit{implicitly} through the algorithm or model design (e.g., auxiliary tasks in multi-task learning, or ordering of individual models in model pipelines).} \gog{Both, we argue, conform to the above operational definition of knowledge to which we subscribe in this work.} \al{Finally, we emphasize that we consider the annotated corpora, leveraged in supervised task learning, to be input and not external knowledge brought to facilitate learning.}

\subsection{Analysis Scope} 

Generally, we focus on natural language argumentation \alfinal{and its computational treatment in NLP. } %
Hence, we exclude work \alfinal{outside of this community, e.g., studies} on abstract argumentation~\citep[e.g.,][]{vreeswijk1997abstract}, \alfinal{except if there is a strong link to natural language argumentation.} \hwfinal{For articles published in non-NLP venues, we made the decision based on the title}. \hwfinal{When unclear from the title whether the work primarily addresses natural language argumentation, e.g., as in the case of \citet{mcburney:2021}, we analyzed the whole article before making the scope decision}. Our survey covers the four subareas of CA \alfinal{in NLP} from \S\ref{sec:background}, with the following restrictions:\,\,

In {\em argument mining}, we do not include methods that have been designed strictly for a specific genre or domain and are not applicable elsewhere. \textit{Argumentative zoning} \citep[e.g.,][]{teufel1999argumentative,teufel2009towards,mo2020deep} and \textit{citation analysis} \citep[e.g.,][]{athar_sentiment_2011,lauscher2021multicite}, both specific to scientific publications, exemplify such methods. In contrast, we include methods that the general mining of argumentative structures, even if evaluated only in specific domains \citep[e.g.,][]{lauscher-etal-2018-investigating}. 

In {\em argument assessment}, we exclude work targeting sentiment analysis~\citep[e.g.,][]{socher-etal-2013-recursive,wachsmuth2014modeling}, \al{as it is inherently more generic than other argumentation tasks and, accordingly, well-explored in general natural language understanding}. Also, we exclude work on general-purpose natural language inference and common-sense reasoning \cite{bowman2015large,rajani2019explain,ponti2020xcopa} in {\em argument reasoning}, and we do not cover the body of work on leveraging external structured knowledge for improved reasoning~\citep[e.g.,][]{forbes-etal-2020-social,lauscher-etal-2020-common}; we view these methods as more generic reasoning approaches that can, among others, also support argumentative reasoning~\citep[e.g.,][]{habernal-etal-2018-argument}, which we do cover in this survey.
Finally, our overview of {\em argument generation} is limited strictly to argumentative text generation, as in argument summarization~\citep[e..g,][]{syed-etal-2020-news} and claim synthesis~\citep[e.g.,][]{bilu-slonim-2016-claim}. The enormous body of work on (non-argumentative) natural language generation \cite{gatt2018survey} is out of our scope. 

\todo{Note that some applications of CA are typically addressed through larger systems, which are composed of models tackling several of the tasks above. For instance,  in \emph{argument search}, a system might be composed of an argument extraction component (\emph{mining}), a retrieval component that determines relevant arguments, as well as a quality rating component (\emph{assessment}) to rank the mined arguments retrieved for given a topic \cite{wachsmuth:2017e}. In this work, we focus on core CA tasks and do not specifically discuss such composite systems.} \alfinal{Within the described scope, we aim for comprehensiveness. However, given the immense body of work on natural language argumentation, we do not claim that this survey is complete.}

\subsection{Analysis and Annotation Process} 

We survey the state of the art in CA through the prism of the knowledge types leveraged in existing approaches. For each of the four CA subareas, we conducted our literature research in two steps: (1) in a {\em pre-study}, we collected \textit{all} papers that we saw as relevant. To this end, we combined our expert knowledge of the field with extensive search in scientific search engines and proceedings of relevant conferences and workshops. On this basis, we established the knowledge pyramid. (2) In an \emph{in-depth study}, we then selected the 10 most representative papers (according to scientometric indicators and our expert judgment) for each subarea and annotated them with the types of knowledge from the pyramid. \todo{We instructed three expert annotators to read each paper carefully. Based on our knowledge definition above and common forms of knowledge we identified in the pre-study, they were asked to decide \emph{what types and what forms} of knowledge were involved, thus assigning all applicable types from the pyramid to each of the 40 sampled papers.}

\paragraph{Agreement.} 

\todo{We measured inter-annotator agreement (IAA) in a \emph{top-level} and an \emph{all-levels} variant across all sampled $40$ papers (10 for each CA area) in terms of pair-wise averaged Cohen's $\kappa$ score. First, for each of the papers, we determined the most specific type of knowledge that it exploits (i.e., the one that is highest in the pyramid). 
Here, we observe a moderate IAA~\citep{landis1977measurement} with $\kappa = 0.54$. Second, across all categories, we observe a substantial IAA of $\kappa=0.74$.} All cases of disagreement were discussed thoroughly and resolved jointly. 

The final distribution of knowledge types identified in papers  for each CA subarea is shown in Figure~\ref{pyramid}b. Expectedly, almost all works (36 out of 40) leverage linguistic knowledge in some form. In contrast, \todo{world and topic knowledge} (e.g., common-sense and factual knowledge, logic and rules) seem to be used least across the board. 
\hw{A reason for the latter may lie in the computational complexity of encoding such knowledge in a way that it can benefit concrete approaches to tasks---whereas this is often much more straightforward for argumentation-specific knowledge (e.g., using lexicons) and task-specific knowledge (e.g., adopting a multitask learning setup). Moreover, topic knowledge is likely to make approaches more topic-dependent,} \gog{i.e., less broadly applicable, which is, more generally, often seen as an undesirable property for NLP approaches.}
We discuss the distribution in detail in the next section.

\paragraph{Pre-Study.}  

Our aim was to collect as many relevant publications as we could for each of the four CA subareas. We first compiled a list of publications that we were personally aware of (i.e., leveraging ``expert knowledge''). Then, we augmented the list by firing queries with relevant keywords (again, compiled based on our expert knowledge) against the ACL Anthology\footnote{\url{https://aclanthology.org/}} and Google Scholar.\footnote{\url{https://scholar.google.com/}} 

For example, we used the following queries for argument mining: \textit{``argument[ation] mining''}, \textit{``argument[ative] component''}, \textit{``argument[ative] relation''}, and \textit{``argument[ative] structure''}. For argument generation, we queried \textit{``argument generation''}, \textit{``argument synthesis''}, \textit{``claim generation''}, \textit{``claim synthesis''}, and \textit{``argument summarization''}. %
In addition, we examined all publications from the proceedings of all seven editions (2014--2020) of the Argument Mining workshop series.

In each subarea, we included only publications that propose a computational approach to solving (at least) one CA task; in contrast, we did not consider publications describing shared tasks~\cite{habernal-etal-2018-argument} or external knowledge resources for CA \cite{alkhatib:2020a}. With these rules in place, we ultimately collected a total of 162 CA papers, entirely listed in Table \ref{tab:allpapers}. By analyzing the types of knowledge used by approaches from collected publications, we induced the pyramid of knowledge types in Figure~\ref{pyramid} with four coarse-grained knowledge types (\S\ref{sec:pyramid}), which was then the basis for our in-depth study (\S\ref{ssec:res_mining}--\S\ref{ssec:res_generation}).

\begin{table*}
\tiny
\begin{tabularx}{\linewidth}{XlX||XlX}
\toprule
    \bf Task & \bf Paper & \bf Top Pyramid Level &  \bf Task & \bf Paper & \bf Top Pyramid Level  \\
\midrule
\multicolumn{6}{c}{\textit{Argument Mining}} \\
 \midrule
Comp. identification 		& \citet{boltuzic-snajder-2014-back}  				& World and topic 	&Multiple tasks 			& \citet{stab-gurevych-2014-identifying} 		& Arg.-specific \\
 					& \textbf{\citet{ajjour-etal-2017-unit}} 				&Linguistic 		&					&  \citet{persing-ng-2020-unsupervised} 		&Task-specific \\
 					& \citet{spliethover-etal-2019-worth} 				& Linguistic 		&					&  \citet{lawrence-reed-2015-combining} 		&Arg.-specific \\
 					& \citet{petasis-2019-segmentation} 				& Linguistic		&					&  \citet{sobhani-etal-2015-argumentation} 	&Arg.-specific \\
 					& \citet{trautmann2020} 						& Linguistic 		&					&  \textbf{\citet{peldszus-stede-2015-joint}} 	&Task-specific \\
Comp. classification		& \citet{ong-etal-2014-ontology} 				& Linguistic 		&					&  \citet{persing-ng-2016-end} 				&Arg.-specific \\
 					& \citet{sobhani-etal-2015-argumentation} 		& Arg.-specific 		&					&  \textbf{\citet{eger-etal-2017-neural}} 		&Linguistic 	\\
 					& \citet{rinott-etal-2015-show} 					& Task-specific 		&					&  \textbf{\citet{lawrence-reed-2017-mining}}	& Arg.-specific 		\\
 					& \citet{al-khatib-etal-2016-cross} 				& Linguistic 		&					&  \citet{lawrence-reed-2017-using} 			& Arg.-specific 		\\
 					& \citet{liebeck-etal-2016-airport} 				& Linguistic 		&					&  \citet{potash-etal-2017-heres} 			& Arg.-specific 		\\
 					& \textbf{\citet{daxenberger-etal-2017-essence}} 	&Linguistic 		&					&  \citet{aker-etal-2017-works} 				&Arg.-specific 		\\
 					& \textbf{\citet{levy-etal-2017-unsupervised}} 		& Arg.-specific 		
 					&					&  \textbf{\citet{niculae-etal-2017-argument}} 	&Arg.-specific 		\\
 					& \citet{shnarch-etal-2017-grasp} 				&Arg.-specific 		&					&  \citet{stab2017parsing} 					& Arg.-specific 		\\
 					& \citet{habernal-gurevych-2017-argumentation} 	&Arg.-specific 		&					&  \citet{saint-dizier-2017-using} 			& Task-specific 		\\
 					& \citet{dusmanu-etal-2017-argument} 			&Arg.-specific 		&					&  \citet{schulz-etal-2018-multi}				& Linguistic 		\\
 					& \citet{lauscher-etal-2018-investigating} 			&Arg.-specific 		&					&  \citet{shnarch-etal-2018-will} 			&Linguistic 		\\
 					& \textbf{\citet{lugini-litman-2018-argument}} 		& Arg.-specific 		&					&  \citet{eger-etal-2018-cross} 				& Linguistic 		\\
 					& \citet{stab-etal-2018-cross} 					& Arg.-specific 		&					&  \citet{morio-fujita-2018-end} 				& Arg.-specific 		\\
 					& \citet{jo-etal-2019-cascade} 					&Linguistic 		&					&  \citet{gemechu-reed-2019-decompositional} 	&Linguistic 		\\
 					& \citet{mensonides-etal-2019-automatic} 			&Arg.-specific 		&					&  \citet{lin-etal-2019-lexicon} 				&Arg.-specific \\
 					& \citet{reimers-etal-2019-classification} 			&Arg.-specific 		&					&  \citet{hewett-etal-2019-utility} 			&Arg.-specific \\
 					& \citet{hua-etal-2019-argument} 				&Arg.-specific 		&					&  \citet{haddadan-etal-2019-yes} 			&Arg.-specific \\
Relation identification 	&\textbf{\citet{cabrio-villata-2012-combining}} 		& World and topic 	&					&  \citet{eide-2019-swedish} 				&Arg.-specific \\
 					& \citet{carstens-toni-2015-towards} 				&Arg.-specific		&					&  \citet{chakrabarty-etal-2019-ampersand} 	&Arg.-specific \\
 					& \citet{cocarascu-toni-2017-identifying} 			&Linguistic 		&					&  \citet{huber-etal-2019-aligning} 			&Arg.-specific \\
 					& \citet{hou-jochim-2017-argument} 				&Task-specific 		&					&  \citet{accuosto-saggion-2019-transferring} 	&Task-specific \\
 					& \textbf{\citet{galassi-etal-2018-argumentative}} 	&Linguistic 		&					&  \citet{morio-etal-2020-towards} 			&Linguistic \\
 					& \citet{paul2020argumentative} 				&World and topic 	&					&  \citet{wang-etal-2020-argumentation} 		&Arg.-specific \\
\midrule
\multicolumn{6}{c}{\textit{Argument Assessment}} \\ 
 \midrule
Stance Detection 		&\citet{ranade-etal-2013-stance} 				&Arg.-specific 		&Quality assessment 	&\textbf{\citet{Habernal2016WhichAI}} 		&Linguistic 		\\
					&\citet{hasan-ng-2014-taking} 					&Linguistic 		&					&\citet{ghosh-etal-2016-coarse} 			&Arg.-specific 		\\
					&\citet{sobhani-etal-2015-argumentation} 			&Arg.-specific 		&					&\citet{wachsmuth-etal-2016-using} 			&Arg.-specific 		\\
					&\citet{persing-ng-2016-modeling} 				&Arg.-specific 		&					&\citet{wei-etal-2016-post} 				&Task-specific 		\\
					&\citet{toledo-ronen-etal-2016-expert} 			&Task-specific		&					&\citet{10.1145/2872427.2883081} 			&Task-specific 		\\
					&\citet{sobhani-etal-2017-dataset} 				&Linguistic 		&					&\citet{chalaguine-schulz-2017-assessing} 	&Linguistic 		\\
					&\textbf{\citet{bar-haim-etal-2017-stance}} 		&Arg.-specific 		&					&\citet{stab-gurevych-2017-recognizing} 		&Linguistic 		\\
					&\citet{boltuzic-snajder-2017-toward} 			&Task-specific 		&					&\citet{potash-etal-2017-length} 			&Linguistic 		\\
					&\citet{bar-haim-etal-2017-improving} 			&Task-specific 		&					&\textbf{\citet{wachsmuth-etal-2017-pagerank}}	&Arg.-specific 		\\
					&\citet{rajendran-etal-2018-something} 			&Linguistic 		&					&\citet{ijcai2017-570} 					&Task-specific 		\\
					&\citet{sun-etal-2018-stance} 					&Arg.-specific 		&					&\citet{lukin-etal-2017-argument} 			&Task-specific 		\\
					&\citet{rajendran-etal-2018-sentiment} 			&Arg.-specific 		&					&\citet{wachsmuth-etal-2017-computational} 	&Task-specific 		\\
					&\citet{kotonya-toni-2019-gradual} 				&Linguistic 		&					&\citet{simpson-gurevych-2018-finding} 		&Linguistic 		\\
					&\citet{durmus-etal-2019-determining} 			&Linguistic 		&					&\citet{gu-etal-2018-incorporating} 			&Linguistic 		\\
					&\citet{durmus-cardie-2019-corpus} 				&Task-specific 		&					&\citet{passon-etal-2018-predicting} 			&Arg.-specific 		\\
					&\citet{toledo-ronen-etal-2020-multilingual} 		&Linguistic 		&					&\citet{ji-etal-2018-incorporating} 			&Task-specific 		\\
					&\citet{kobbe-etal-2020-unsupervised} 			&Arg.-specific 		&					&\textbf{\citet{durmus-cardie-2018-exploring}} 	&Task-specific 		\\
					&\citet{sirrianni-etal-2020-agreement} 			&Arg.-specific 		&					&\citet{el-baff-etal-2018-challenge} 			&Task-specific 		\\
					&\citet{somasundaran-wiebe-2010-recognizing}		&Arg.-specific 		&					&\citet{10.1145/3331184.3331282} 			&Linguistic 		\\
					&\citet{porco-goldwasser-2020-predicting} 		&Task-specific 		&					&\citet{10.1145/3331184.3331327} 			&Linguistic 		\\
					&\citet{scialom-etal-2020-toward} 				&Task-specific 		&					&\citet{gleize-etal-2019-convinced} 			&Linguistic 		\\
Frame identification		&\citet{ajjour-etal-2019-modeling} 				&Task-specific 		&					&\citet{toledo-etal-2019-automatic} 			&Linguistic 		\\
					&\textbf{\citet{trautmann-2020-aspect}} 			&Linguistic 		&					& \citet{potash-etal-2019-ranking} 			&Linguistic 		\\
Quality assessment 		&\citet{4781139} 							&Task-specific 		&					& \textbf{\citet{gretz2020large}} 			&Linguistic 		\\
					&\citet{persing-etal-2010-modeling} 				&Linguistic 		&					& \textbf{\citet{el-baff-etal-2020-analyzing}} 	&Linguistic 		\\
					&\citet{persing-ng-2013-clarity} 					&Linguistic 		&					& \citet{wachsmuth-werner-2020-intrinsic} 	&Linguistic 		\\
					&\citet{ong-etal-2014-ontology} 					&Linguistic 		&					& \citet{li-etal-2020-exploring} 				&Arg.-specific 		\\
					&\citet{persing-ng-2014-modeling} 				&Linguistic 		&					& \textbf{\citet{al-khatib-etal-2020-exploiting}} 	&Task-specific		\\
					&\citet{song-etal-2014-applying} 				&Arg.-specific 		&					& \citet{lauscher-etal-2020-rhetoric} 			&Task-specific 		\\
					&\textbf{\citet{persing-ng-2015-modeling}} 		&Arg.-specific 		&					& \citet{skitalinskaya-etal-2021-learning} 		&Linguistic 		\\
					&\citet{stab-gurevych-2016-recognizing} 			&Linguistic 		&Other tasks			&\textbf{\citet{kobbe-etal-2020-exploring}} 	&Task-specific 		\\
					&\citet{habernal-gurevych-2016-makes} 			&Linguistic 		&					&\citet{yang-etal-2019-lets} 				&Linguistic 		\\
\midrule
\multicolumn{6}{c}{\textit{Argument Reasoning}} \\ 
\midrule
Warrant identification 	&\textbf{\citet{boltuzic-snajder-2016-fill}} 			& Linguistic		& Scheme classification	& \textbf{\citet{feng-hirst-2011-classifying}} 		&Task-specific 		 \\
					&\citet{sui-etal-2018-joker} 					& Linguistic		&					& \citet{song-etal-2014-applying} 				&Task-specific 		 \\
					&\citet{liebeck-etal-2018-hhu} 					&Linguistic		&					& \textbf{\citet{lawrence-reed-2015-combining}}		&Linguistic 		\\
					&\textbf{\citet{tian-etal-2018-ecnu}} 				&Linguistic		&					& \textbf{\citet{liga-2019-argumentative}} 			&Linguistic 		 \\
					&\citet{brassard-etal-2018-takelab} 				&Linguistic		&					& %
					\\
					& \citet{sui-etal-2018-joker} 					&Linguistic		& Fallacy Recognition 	& \textbf{\citet{habernal-etal-2018-name}} 			&Linguistic 		 \\
					&\textbf{\citet{botschen-etal-2018-frame}} 			&World and topic	&					&\citet{HABERNAL18.494} 					&Linguistic 		\\
					&\textbf{\citet{choi-lee-2018-gist}} 				&World and topic	&					&\textbf{\citet{delobelle-etal-2019-computational}} 	&Linguistic 		\\
					&\textbf{\citet{niven-kao-2019-probing}} 			&World and topic 	& Other tasks			&\citet{becker-etal-2021-reconstructing}			& World and topic \\
\midrule
\multicolumn{6}{c}{\textit{Argument Generation}} \\ 
\midrule
Summarization 			& \citet{egan-etal-2016-summarising} 			&Linguistic 		&Argument synthesis 	&\textbf{\citet{zukerman-etal-2000-using}} 	&Task-specific \\
					& \textbf{\citet{wang-ling-2016-neural}} 			&Linguistic 		&					&\citet{CARENINI2006925} 				&Task-specific\\
					& \citet{syed-etal-2020-news} 					&Arg.-specific 		&					&\textbf{\citet{sato-etal-2015-end}} 			&Linguistic \\
					& \citet{10.1145/3397271.3401186} 				&Arg.-specific 		&					&\citet{reisert-etal-2015-computational} 		&Arg.-specific \\
					& \textbf{\citet{bar-haim-etal-2020-quantitative}}		&Arg.-specific 		&					&\citet{hua-wang-2018-neural} 				&World and topic \\
Claim Synthesis 		&\textbf{\citet{bilu-slonim-2016-claim}} 			&Task-specific 		&					&\citet{wachsmuth-etal-2018-argumentation} 	&Arg.-specific \\
					&\citet{chen-etal-2018-learning} 				&World and topic 	&					&\citet{le-etal-2018-dave} 					&Arg.-specific 		\\
					&\citet{hidey-mckeown-2019-fixed} 				&Arg.-specific 		&					&\citet{hua-etal-2019-argument-generation} 	&World and topic \\
					&\citet{alshomary-etal-2020-target}				&Arg.-specific 		&					&\textbf{\citet{hua-wang-2019-sentence}} 		&World and topic \\
					&\textbf{\citet{gretz-etal-2020-workweek}} 			&Arg.-specific		&					&\textbf{\citet{el-baff-etal-2019-computational}} 	&Arg.-specific  \\
					&\textbf{\citet{alshomary-etal-2021-belief}} 		&Task-specific		&					&\citet{bilu-etal-2019-argument} 			&Task-specific \\
					&										&				&					&\textbf{\citet{schiller-etal-2021-aspect}} 		&Task-specific \\
\bottomrule
\end{tabularx}
\caption{List of all publications surveyed in this study, across the four subareas of CA (argument mining, argument assessment, argument reasoning, and argument generation) sorted by task and year of publication, with the indication of the most specific level of knowledge used. Publications in bold are those selected for the in-depth study (\S\ref{sec:methodology}).}
\label{tab:allpapers}
\end{table*}

\paragraph{In-Depth Study.} 

In the second step, we used the knowledge pyramid as the basis for an in-depth analysis of a subset of 40 publications (10 per research area; bold in Table \ref{tab:allpapers}). Our selection of \textit{prominent} papers for the in-depth study was guided by the following set of (sometimes mutually conflicting) criteria: (1) maximize the scientific impact of the publications in the sample, measured as a combination of the number of publication citations and our expert judgment of publication's overall impact on the CA field or subarea; (2) maximize the number of different methodological approaches in the sample;\footnote{Note that diversifying the sample with respect to methods is different than diversifying it according to knowledge types: two approaches may use the same type(s) of knowledge (e.g., linguistic) while adopting different methods (e.g., syntactic features vs. neural LMs). Our aim was to reduce the methodological redundancy of the sample.} and (3) maximize the representation of different researchers and research groups. 

Once we had selected the 40 publications, three authors of this manuscript independently labeled all of them with the knowledge types from the pyramid. This allowed us to measure the  inter-annotator agreement and to test the extent of shared understanding of the knowledge types captured by the pyramid and their usage in individual methodological approaches in CA.          
While we are aware that we cannot draw statistically significant conclusions based on a sample of such a limited size, we believe that our findings and this in-depth perspective will still be informative for the CA community.

\section{Knowledge in Argumentation}
\label{sec:taxonomy}
As a result of our survey, we now introduce the \textit{argumentation knowledge pyramid}, our proposed taxonomy encompassing four coarse-grained types of knowledge leveraged in CA. We then profile the large body of papers from the four CA subareas through the lens of the pyramid. 

\hwfigure{pyramid}{(a)~Our proposed \textit{argumentation knowledge pyramid}, encompassing four coarse-grained types of knowledge leveraged in CA research: the pyramid is organized according to increasing specificity of the knowledge types, from the bottom to the top. (b)~Relative frequencies of the presence of knowledge types in the 40 representative papers (10 per CA subarea: mining, assessment, reasoning, and generation) selected for our in-depth study.
}

\subsection{Argumentation Knowledge Pyramid}
\label{sec:pyramid}

Based on the findings of our pre-study, we identify four coarse-grained types of knowledge being leveraged in computational argumentation research, which we organize in a taxonomy, as depicted in Figure~\ref{pyramid}. We chose to visualize our organization as a pyramid because it allows us to express a hierarchical generality-specificity relationship between the different types of knowledge.

\paragraph{Linguistic Knowledge.} 

At the bottom of the pyramid is the linguistic knowledge, leveraged by virtually all CA models and needed in practically all NLP tasks. In our pyramid, linguistic knowledge is a broad category that includes features derived from word {\em n}-grams, information about linguistic structure (e.g., part-of-speech tags, dependency parses), as well as features based on models of distributional semantics, such as (pre-trained) word embedding spaces \citep[e.g.,][]{mikolov2013distributed,pennington-etal-2014-glove,bojanowski2017enriching} or representation spaces spanned by neural language models (LMs) \citep[e.g.,][]{clark2020electra,devlin-etal-2019-bert}. We also consider leveraging distributional spaces (word embeddings or pretrained LMs) built for specific (argumentative) tasks and domains as a form of linguistic knowledge, since such representation spaces are induced purely from textual corpora without any external supervision signal.

\paragraph{World and Topic Knowledge.} 

Above the linguistic knowledge, we place the category of \textit{world and topic knowledge} in which we bundle all types of knowledge that are generally considered useful for various natural language understanding tasks, but that are not (or even cannot be) directly derived from textual corpora. 
This includes all types of common-sense knowledge, task-independent world knowledge (also known as factual knowledge), logical general-purpose axioms and rules, and similar. In most cases, such knowledge is collected from external structured or semi-structured resources \cite{sap2020commonsense,lauscher-etal-2020-common,ji2021survey}. \todo{Knowledge about a specific \emph{debate topic} (e.g., legalization of marihuana) falls under this category, since topics encompass a set of real-world concepts (e.g., marihuana) and related facts (e.g., medical aspects of marihuana usage). Some systems explicitly require the debate topic as input, in order to gather topc knowledge from external sources.}

\paragraph{Argumentation-Specific Knowledge.}

The third category in our knowledge pyramid encompasses \todo{knowledge abour \textit{what constitutes argumentation, arguments, and argumentative language}, including knowledge about subjective language~\citep{stede2018argumentation}.}
This includes models of argumentation and argumentative structures \cite{toulmin_uses_2003,bentahar2010taxonomy}, models of cultural aspects and moral values \cite{haidt2004intuitive,graham2013moral}, lexicons with terms indicating subjective, psychological, and moral categories \cite{hu2004mining,tausczik2010psychological,graham2009liberals}, predictions of subjectivity and sentiment classification models \cite{socher-etal-2013-recursive}, etc. \gog{While sentiment, emotions, and affect are not argumentative \textit{per se}, subjectivity is ingrained in argumentation and strongly influences argumentative manifestations (or lack thereof).}

\paragraph{Task-Specific Knowledge.} 

As the most specific type of knowledge, this category covers the types of knowledge that are relevant only for a specific CA task or a small set of tasks. For instance, leveraging discourse structure is considered beneficial for argumentative relation identification \cite{stab-gurevych-2014-identifying,persing-ng-2016-end,opitz-frank-2019-dissecting}, a common argument mining task.

\begin{table*}[t!]
\def\arraystretch{0.9}
\scriptsize
\begin{tabularx}{\linewidth}{l l l l l}
\toprule
\bf Knowledge & \bf Source & \bf CA Subarea (Task) & \bf Introduced & \bf Explanation \\ 
\midrule
\specialcell{\parbox[t]{1.05cm}{Linguistic}} &
\specialcell{\parbox[t]{1.7cm}{\newcite{habernal-etal-2018-name}}} &
\specialcell{\parbox[t]{2.45cm}{Argument reasoning \\ (fallacy recognition)}} &
\gog{Explicitly} &
\specialcell{\parbox[t]{7.3cm}{Semantic associations between lexical units in the word embedding space enable generalization across different \textit{lexicalizations} of ad hominem arguments (e.g., \textit{``pretentions [explanation]''} vs. \textit{``narcissistic [idiot]''}) and wordings that point to fallacious reasoning (e.g., \textit{``[if only you wouldn't rely on] fallacious arguments''} vs. \textit{``[another] unsubstantiated statement)''}}}. \\ 
\midrule
\specialcell{\parbox[t]{1.05cm}{World \\ and topic}} &
\specialcell{\parbox[t]{1.7cm}{\newcite{hua-wang-2019-sentence}}} &
\specialcell{\parbox[t]{2.45cm}{Argument generation \\ (argument synthesis)}} &
\gog{Implicitly} &
\specialcell{\parbox[t]{7.3cm}{The structure of the argument -- sequence of \textit{Premise}, \textit{Claim}, and \textit{Functional} utterances -- is conditioned by the \textit{topic} of debate. E.g., Reddit arguments in political topics (e.g., \textit{``US cutting off foreign aid''} tend to start with a \textit{Claim} (\textit{``It can be a useful
political bargaining chip''}), continue with supporting \textit{Premises} (e.g., \textit{``US cut financial aid to Uganda due to its plans to make homosexuality a crime''}) and finish with \textit{Functional} utterances (e.g., \textit{``Please change your mind!''}).}} \\ 
\midrule
\specialcell{\parbox[t]{1.05cm}{Arg.- \\ specific}} & 
\specialcell{\parbox[t]{1.7cm}{\newcite{wachsmuth-etal-2017-pagerank}}} & 
\specialcell{\parbox[t]{2.45cm}{Argument assessment \\ (quality assessment)}} &
\gog{Explicitly} &
\specialcell{\parbox[t]{7.3cm}{\textit{Argument relevance} is determined in an ``objective'' way. Argument ``reuse'', where one argument leverages the conclusion of another argument is the base for the induction of a large-scale (directed) argument graph. Running a PageRank algorithm on that graphs yields relevance scores for all arguments. Such objective and content-agnostic argument relevance score can be \textit{useful for a wide variety of CA tasks}; knowledge about argument reuse thus represents argumentation-specific knowledge.}} \\
\midrule
\specialcell{\parbox[t]{1.05cm}{Task- \\ specific}} & 
\specialcell{\parbox[t]{1.7cm}{\newcite{peldszus-stede-2015-joint}}}  & 
\specialcell{\parbox[t]{2.45cm}{Argument mining \\ (multiple tasks)}} & 
\gog{Implicitly} & 
\specialcell{\parbox[t]{7.3cm}{Argumentative structure of the text assumed to be a \textit{tree}: there is one central claim for the text which is the root of the tree, other argumentative components are the nodes of the tree, and edges reflect the \textit{support} or \textit{attack} relations between argumentative discourse components.}} \\ 
\bottomrule
\end{tabularx}
\caption{Concrete examples for the four types of knowledge distinguished in the knowledge pyramid (see Figure~\ref{pyramid}). \gog{We additionally indicate for each example whether the knowledge is introduced in an explicit or implicit manner (column ``Introduced''; see \S\ref{sec:knowledge})}}
\label{tab:know_type_examples}
\vspace{-1em}
\end{table*}

\begin{table}[t!]
\tiny
\renewcommand{\arraystretch}{1}
\setlength{\tabcolsep}{5pt}%
\begin{tabular}{llllll}
\toprule
\bf Approach       	& \bf Linguistic	& \bf World 	& \bf Arg. 		& \bf Task. \\
				&			& \bf and Topic	& \bf specific	& \bf Specific \\
\midrule
\multicolumn{5}{c}{\textit{Argument Mining}} \\ 
\midrule	
\citet{cabrio-villata-2012-combining} &      {\color{lightgray}\xmark} &  \cmark &                 {\color{lightgray}\xmark} &        {\color{lightgray}\xmark} \\
\citet{peldszus-stede-2015-joint} &           \cmark &  {\color{lightgray}\xmark} &                 \cmark &        \cmark \\
     \citet{daxenberger-etal-2017-essence}  &     \cmark &  {\color{lightgray}\xmark} &                 {\color{lightgray}\xmark} &        {\color{lightgray}\xmark} \\
\citet{eger-etal-2017-neural} &     \cmark &  {\color{lightgray}\xmark} &                 {\color{lightgray}\xmark} &        \cmark \\
\citet{niculae-etal-2017-argument}  &     \cmark &  {\color{lightgray}\xmark} &                 {\color{lightgray}\xmark} &        {\color{lightgray}\xmark} \\
 \citet{lawrence-reed-2017-using}  &     \cmark &  \cmark &                 \cmark &        {\color{lightgray}\xmark} \\
\citet{levy-etal-2017-unsupervised}   &     \cmark &  \cmark &                \cmark
&        {\color{lightgray}\xmark} \\
\citet{ajjour-etal-2017-unit}   &     \cmark &  {\color{lightgray}\xmark} &                 \cmark &        {\color{lightgray}\xmark} \\
\citet{galassi-etal-2018-argumentative}  &     \cmark &  {\color{lightgray}\xmark} &                 {\color{lightgray}\xmark} &        {\color{lightgray}\xmark} \\
\citet{lugini-litman-2018-argument}  &     \cmark &  {\color{lightgray}\xmark} &                 {\color{lightgray}\xmark} &        \cmark \\
\midrule
\multicolumn{5}{c}{\textit{Argument Assessment}} \\ 
\midrule
\citet{persing-ng-2015-modeling} &       \cmark &  {\color{lightgray}\xmark} &                 \cmark &        {\color{lightgray}\xmark} \\
\citet{Habernal2016WhichAI}  &     \cmark &  {\color{lightgray}\xmark} &                 \cmark &        {\color{lightgray}\xmark} \\
\citet{wachsmuth-etal-2017-pagerank}  & {\color{lightgray}\xmark} &  {\color{lightgray}\xmark} &                 \cmark &        {\color{lightgray}\xmark} \\
\citet{bar-haim-etal-2017-stance}  &     \cmark &  \cmark &                 \cmark &        {\color{lightgray}\xmark} \\
\citet{durmus-cardie-2018-exploring}  &     \cmark &  {\color{lightgray}\xmark} &                 \cmark &        \cmark \\  
\citet{trautmann-2020-aspect}  &     \cmark &  {\color{lightgray}\xmark} &                 {\color{lightgray}\xmark} &        {\color{lightgray}\xmark} \\
\citet{kobbe-etal-2020-exploring}  &     \cmark &  {\color{lightgray}\xmark} &                 \cmark &        {\color{lightgray}\xmark} \\
\citet{el-baff-etal-2020-analyzing}  &     \cmark &  {\color{lightgray}\xmark} &                 \cmark &        \cmark \\
\citet{al-khatib-etal-2020-exploiting}  &     \cmark &  {\color{lightgray}\xmark} &                 {\color{lightgray}\xmark} &        \cmark \\
\citet{gretz2020large}  &     \cmark &  {\color{lightgray}\xmark} &                 {\color{lightgray}\xmark} &        {\color{lightgray}\xmark} \\  
\midrule
\multicolumn{5}{c}{\textit{Argument Reasoning}} \\ 
\midrule
\citet{feng-hirst-2011-classifying}  & {\color{lightgray}\xmark} &  {\color{lightgray}\xmark} &                 {\color{lightgray}\xmark} &        \cmark \\  
\citet{lawrence-reed-2015-combining}  &     \cmark &  {\color{lightgray}\xmark} &                 {\color{lightgray}\xmark} &        \cmark \\   
\citet{boltuzic-snajder-2016-fill}  &     \cmark &  {\color{lightgray}\xmark} &                 {\color{lightgray}\xmark} &        {\color{lightgray}\xmark} \\
\citet{habernal-etal-2018-name}  &     \cmark &  {\color{lightgray}\xmark} &                 {\color{lightgray}\xmark} &        {\color{lightgray}\xmark} \\
\citet{choi-lee-2018-gist}  &     \cmark &  \cmark &                 {\color{lightgray}\xmark} &        {\color{lightgray}\xmark} \\
\citet{tian-etal-2018-ecnu}  &     \cmark &  {\color{lightgray}\xmark} &                 {\color{lightgray}\xmark} &        {\color{lightgray}\xmark} \\
\citet{botschen-etal-2018-frame}  &     \cmark &  \cmark &                 {\color{lightgray}\xmark} &        {\color{lightgray}\xmark} \\
\citet{delobelle-etal-2019-computational}  &     \cmark &  {\color{lightgray}\xmark} &                 {\color{lightgray}\xmark} &        {\color{lightgray}\xmark} \\
\citet{niven-kao-2019-probing}  &     \cmark &  {\color{lightgray}\xmark} &                 {\color{lightgray}\xmark} &        {\color{lightgray}\xmark} \\
\citet{liga-2019-argumentative}  &     \cmark &  {\color{lightgray}\xmark} &                 {\color{lightgray}\xmark} &        {\color{lightgray}\xmark} \\  
\midrule
\multicolumn{5}{c}{\textit{Argument Generation}} \\ 
\midrule
\citet{zukerman-etal-2000-using}  &     {\color{lightgray}\xmark} &  {\color{lightgray}\xmark} &                 \cmark &        \cmark \\
\citet{sato-etal-2015-end}  &     \cmark &  {\color{lightgray}\xmark} &                 \cmark &        {\color{lightgray}\xmark} \\
\citet{bilu-slonim-2016-claim}  &     \cmark &  {\color{lightgray}\xmark} &                 \cmark &        \cmark \\
\citet{wang-ling-2016-neural}  &     \cmark &  {\color{lightgray}\xmark} &                 {\color{lightgray}\xmark} &        {\color{lightgray}\xmark} \\
\citet{el-baff-etal-2019-computational}  &     \cmark &  {\color{lightgray}\xmark} &                 {\color{lightgray}\xmark} &        \cmark \\
\citet{hua-etal-2019-argument}  &     \cmark &  \cmark &                 {\color{lightgray}\xmark} &        {\color{lightgray}\xmark} \\
\citet{bar-haim-etal-2020-quantitative}  &     \cmark &  {\color{lightgray}\xmark} &                 \cmark &        {\color{lightgray}\xmark} \\
\citet{gretz-etal-2020-workweek}  &     \cmark &  {\color{lightgray}\xmark} &                 {\color{lightgray}\xmark} &        {\color{lightgray}\xmark} \\
\citet{alshomary-etal-2021-belief}  &     \cmark &  {\color{lightgray}\xmark} &                 {\color{lightgray}\xmark} &        \cmark \\
\citet{schiller-etal-2021-aspect}  &     \cmark &  {\color{lightgray}\xmark} &                 \cmark &        \cmark \\
\bottomrule
\end{tabular}
\caption{The types of knowledge involved in the approaches of all publications that we included in the second stage of our literature survey (in-depth study) ordered by the high-level task they tackle and the year.}
\label{tab:in-depth}
\vspace{-1em}
\end{table}

\medskip \noindent
Table \ref{tab:know_type_examples} illustrates the four types of knowledge from the pyramid by means of concrete examples.

\subsection{Knowledge in {Argument Mining}}
\label{ssec:res_mining}

\paragraph{Pre-Study.} 

From the 162 papers we surveyed, 56 belong to the subarea of argument mining, which is the second-largest subarea after argument assessment. The publications that we analyzed were published in the period from 2012 to 2020. Of these 56 publications, \al{17} relied purely on linguistic knowledge, three exploited world and topic knowledge as the most specific knowledge type, \al{30} leveraged argumentation-specific knowledge, and six task-specific knowledge. We next describe the detailed findings of our in-depth analysis.

\paragraph{In-Depth Study.} 

\todo{Table~\ref{tab:in-depth} shows the results of our assignment of all applicable knowledge types to 10 sampled argument mining papers, published between 2012 and 2018.}
All but one rely on linguistic knowledge: earlier approaches leveraged traditional linguistic features, such as $n$-grams and syntactic features \citep[e.g.,][]{peldszus-stede-2015-joint,lugini-litman-2018-argument}, whereas later work resorted to word embeddings as the dominant representation~\citep[e.g.,][]{eger-etal-2017-neural, niculae-etal-2017-argument, daxenberger-etal-2017-essence, galassi-etal-2018-argumentative}. 

A few papers exploit other types of knowledge. \citet{cabrio-villata-2012-combining}, for example, leverage a pretrained NLI model to analyze online debate interactions.\footnote{Note that our judgments reflect only the types of knowledge that the approach presented in the paper directly exploits: this is why, for example, we judge the reliance of the approach of \citet{cabrio-villata-2012-combining} on a pretrained NLI model as exploitation of world and topic knowledge only, even though the NLI model itself \cite{kouylekov2010open} had been trained using a range of linguistic features.} While they resort to the abstract argumentation framework of \citet{dung1995acceptability}, they do so only for the purposes of the evaluation, which is why we do not judge their approach as reliant on argumentation-specific knowledge.   
\citet{lawrence-reed-2017-using} use, in addition to word embeddings, \todo{world and topic} knowledge from WordNet and argumentation-specific knowledge in the form of structural assumptions for mining large-scale debates. \citet{ajjour-etal-2017-unit} combine linguistic knowledge in the form of GloVe embeddings~\citep{pennington-etal-2014-glove} and other linguistic features with an argumentation-specific lexicon of discourse markers. 
Task-specific mining knowledge is mostly leveraged in multi-task learning scenarios~\citep{lugini-litman-2018-argument} or when aiming to extract arguments of more complex structures, that is, with multiple components and/or chains of claims \citep{eger-etal-2017-neural,peldszus-stede-2015-joint}. For instance, \citet{peldszus-stede-2015-joint} jointly predict different aspects of the argument structure and then apply minimum spanning tree decoding, exploiting that mining of argument structure bears similarities with discourse parsing. 
The only template-based approach we cover is that of \citet{levy-etal-2017-unsupervised}, who construct queries using templates and use ground sentences in Wikipedia concepts (i.e., world and topic knowledge) for unsupervised claim detection. \hw{Their approach also leverages an argumentation-specific lexicon of claim-related words} \gog{(i.e., arg.-specific knowledge), next to the linguistic and world/topic knowledge.}

\subsection{Knowledge in {Argument Assessment}}
\label{ssec:res_assessment}

\paragraph{Pre-Study.} 

The largest portion of the 162 publications, \al{64} in total, belong to the area of argument assessment, spanning the time period from 2008 to 2021. Of those publications, \al{29} leverage only linguistic knowledge, but almost 20 rely on task-specific knowledge as the most specific knowledge type. Interestingly, none of the surveyed papers use \todo{world and topic} knowledge as the most specific knowledge type. That is, if they rely on \todo{world and topic} knowledge, they also leverage argumentation-specific and/or task-specific knowledge.

\paragraph{In-Depth Study.} 

The 10 assessment papers analyzed in-depth (period 2015--2020) reveal that, much like in argument mining, most of the work models linguistic knowledge~\citep[e.g.,][]{trautmann-2020-aspect, kobbe-etal-2020-exploring}. For example, \citet{gretz2020large} assess argument quality based on a representation that combines bag-of-words (i.e., sparse symbolic text representation) with latent embeddings, both derived from static GloVe word embeddings \cite{pennington-etal-2014-glove} and produced by a pretrained BERT model~\citep{devlin-etal-2019-bert}. Most of the papers at the linguistic knowledge level of the pyramid, however, predominantly rely on sparse symbolic (i.e., word-based) linguistic features~\citep[e.g.,][]{persing-ng-2015-modeling, bar-haim-etal-2017-improving,durmus-cardie-2018-exploring,al-khatib-etal-2020-exploiting,el-baff-etal-2020-analyzing}.   

Only one of the 10 selected publications resorts to \todo{world and topic} knowledge: \citet{bar-haim-etal-2017-stance} map the content of claims to Wikipedia concepts for stance classification. A common technique in argument assessment is to include argumentation-specific knowledge about sentiment or subjectivity: this is motivated by the intuition that these features directly affect argumentation quality and correlate with stances. For instance, \citet{wachsmuth-etal-2017-computational} note that \emph{emotional appeal}, which is clearly correlated with the sentiment of the text, may affect the rhetorical effectiveness of arguments. Technically, the information on subjectivity is introduced either by means of subjective lexica~\citep[e.g.,][]{bar-haim-etal-2017-stance,durmus-cardie-2018-exploring,el-baff-etal-2020-analyzing} or via predictions of pretrained sentiment classifiers~\citep[][]{Habernal2016WhichAI}. In a different example of the use of argumentation-specific knowledge, \citet{wachsmuth-etal-2017-pagerank} exploit reuses between arguments (e.g., a premise of one argument uses the claim of another) to quantify argument relevance by means of graph-based propagation with PageRank. 

A notable task-specific knowledge category is the use of \textit{user information} for argument quality assessment. According to theory \citep{wachsmuth-etal-2017-computational}, argument quality does not only depend on the text utterance itself but also on the speaker and the audience, for example, on their prior beliefs and their cultural context. To model this, \citet{durmus-cardie-2018-exploring} include information about users' prior beliefs as predictors of arguments' persuasiveness, \citet{al-khatib-etal-2020-exploiting} predict persuasiveness using user-specific feature vectors, and \citet{el-baff-etal-2020-analyzing} train audience-specific classifiers.

\subsection{Knowledge in {Argument Reasoning}}
\label{ssec:res_reasoning}

\paragraph{Pre-Study.} 

According to our pre-study, argument reasoning is the smallest subarea of CA, with only \al{17} (out of 162) papers published (in the period between 2011 and \al{2021}). The tasks in this subarea include argumentation scheme classification \cite{feng-hirst-2011-classifying,lawrence-reed-2015-combining}, warrant identification and exploitation \cite{habernal-etal-2018-argument,boltuzic-snajder-2016-fill}, and fallacy recognition \cite{habernal-etal-2018-name,delobelle-etal-2019-computational}. Linguistic knowledge denotes the most commonly used type of knowledge in reasoning as well (\al{11 out of 17} papers rely on some type of linguistic knowledge), and four papers in this subarea exploit \todo{world and topic} knowledge.

\paragraph{In-Depth Study.} 

In our subset from argument reasoning, general-domain embeddings are by far the most frequently employed type of knowledge injection approach~\citep{boltuzic-snajder-2016-fill,habernal-etal-2018-name,choi-lee-2018-gist,tian-etal-2018-ecnu,botschen-etal-2018-frame,delobelle-etal-2019-computational,niven-kao-2019-probing}. In contrast, \citet{lawrence-reed-2015-combining} use traditional linguistic features, and \citet{liga-2019-argumentative} models syntactic features with tree kernels to recognize specific reasoning structures in arguments. Task-specific knowledge is modeled by \citet{feng-hirst-2011-classifying}, who design specific features for classifying argumentation schemes, and \citet{lawrence-reed-2015-combining} utilize features specific to individual types of premises and conclusions. \citet{choi-lee-2018-gist} use a pretrained natural language inference model to select the correct warrant in warrant identification.\footnote{Like in the case of \cite{cabrio-villata-2012-combining} in argument mining, we consider a pretrained NLI model to represent \textit{\todo{world and topic} knowledge}.} 
For the same task, \citet{botschen-etal-2018-frame} leverage event knowledge about common situations (from FrameNet) and factual knowledge about entities (from Wikidata).

\subsection{Knowledge in {Argument Generation}}
\label{ssec:res_generation}  

\paragraph{Pre-Study.} 

Finally, we surveyed 23 generation papers, ranging from 2000 to 2021. Argumentation-specific knowledge is the most specific knowledge type in most (10) publications. Six publications have task-specific knowledge as the most specific knowledge type, and four do not employ anything more specific than \todo{world and topic} knowledge. Unlike in other subareas, only few publications (3) in argument generation rely purely on linguistic knowledge. Common argument generation tasks include argument summarization \cite{egan-etal-2016-summarising,bar-haim-etal-2020-quantitative}, claim synthesis \cite{bilu-etal-2019-argument,alshomary-etal-2021-belief}, and argument synthesis \cite{zukerman-etal-2000-using,sato-etal-2015-end}.

\paragraph{In-Depth Study.} 

As in the case of argument reasoning, many generation approaches employ linguistic knowledge in the form of general-purpose embeddings~\citep{wang-ling-2016-neural,hua-etal-2019-argument-generation,bar-haim-etal-2020-quantitative,gretz-etal-2020-workweek,schiller-etal-2021-aspect}. Only~\citet{sato-etal-2015-end} report using traditional (i.e., sparse, symbolic) linguistic features; \citet{bilu-slonim-2016-claim} used traditional linguistic features for predicting the suitability of candidate claims. 

World and topic knowledge is utilized by \citet{hua-etal-2019-argument-generation} who retrieve Wikipedia passages as claim candidates. As argumentation-specific knowledge, \citet{bar-haim-etal-2020-quantitative} use an external quality classifier. In a similar vein, \citet{schiller-etal-2021-aspect} incorporate the output from argument and stance  classifiers from the ArgumenText API \cite{stab2018argumentext} and condition the generation model on control codes encoding topic, stance, and aspect of the argument.
\citet{alshomary-etal-2021-belief} condition their model on a audience beliefs by deriving bag-of-words representations from the authors' texts and then fine-tuning a pretrained language model. \citet{sato-etal-2015-end} model (argumentation-specific) knowledge about values. Predicate and sentiment lexica are employed by \citet{bilu-slonim-2016-claim}, whereas \citet{el-baff-etal-2019-computational} learn likely sequences of argumentative units from features computed from argumentation-specific knowledge. They additionally include task-specific knowledge by using a knowledge base with components of claims. 
A pioneering work that stands out is the approach of \citet{zukerman-etal-2000-using} which uses argumentation-specific knowledge about micro-structure in combination with task-specific discourse templates.

\section{Emerging Trends and Discussion}%
\label{sec:trends}
\begin{table*}[t]
    \centering
    \small
	\renewcommand{\arraystretch}{1}
	\setlength{\tabcolsep}{3.5pt}%
	\begin{tabular}{ll}
    \toprule
    \textbf{Type} & \textbf{Common Modeling Techniques} \\
    \midrule
    Task-specific & Structure (e.g., multitask learning), user information (e.g., features), ... \\
    Argumentation-specific  & Sentiment (e.g., lexicon, external classifier), argumentation (e.g., fine-tuning), ... \\
    \todo{World and topic}   & Inference knowledge (e.g., infusion), world knowledge (e.g., linking to Wikipedia), ... \\
    Linguistic  & $n$-grams (e.g., traditional features), general semantics (e.g., GloVe embeddings), ... \\
    \bottomrule
    \end{tabular}
    \caption{\todo{Common techniques used for modeling the four types of knowledge from the proposed knowledge pyramid.}}
    \label{tab:types_examples}
    \vspace{-1em}
\end{table*}

We now summarize the emerging trends and open challenges in the four CA areas, abstracted from our analyses of the use of knowledge types.

\paragraph{General Observations.} 

Most of the 162 publications that we reviewed aim to capture some type of ``advanced'' knowledge, that is, knowledge beyond what can be inferred from the text data alone: \al{60} publications rely purely on linguistic knowledge, whereas the remaining 102 model at least one of the other three higher knowledge types. This empirically confirms the intuition that success in CA crucially depends on complex knowledge that is external to the text. Also, unsurprisingly, argumentation-specific knowledge is overall the most common type of external knowledge used in CA approaches: \al{argumentation-specific knowledge can, in principle, facilitate any computational argumentation task}. In comparison, world and common-sense knowledge are fairly underrepresented: only \al{seven} of the 40 publications in our in-depth study rely on some variant of it. This is surprising, given that the approaches that leverage such knowledge consistently report substantial performance gains. %

\paragraph{Comparison across Types of Knowledge.}

We observe differences in the form in which the different knowledge types (e.g., linguistic vs.\ argument-specific knowledge) are commonly provided and incorporated in methodological approaches. We provide examples in Table~\ref{tab:types_examples}.

\paragraph{Comparison across Areas.}

We also note substantial differences across the four high-level CA subareas. The predominant most specific knowledge types vary across the areas: in argument mining and assessment, linguistic and argumentation-specific knowledge are most commonly employed, whereas in argument reasoning approaches, \todo{world and topic} knowledge (e.g., knowledge about reasoning mechanisms) represents the most common top-level category from the pyramid. In argument generation, argumentation-specific and task-specific knowledge were the most common top-level categories. We believe that this variance is due to the nature of the tasks in each area: predicting argumentative structures in argument mining is strongly driven by lexical cues (linguistic knowledge) and structural aspects (argumentation-specific knowledge). %
Despite being studied most extensively, argument mining rarely exploits \todo{world and topic}  knowledge (e.g., from knowledge bases or lexico-semantic resources): there is possibly room for progress in argument mining from more extensive exploitation of structured knowledge sources. %

As previously suggested by \citet{wachsmuth-etal-2017-computational}, we find that argument assessment relies on a combination of linguistic features and higher-level argumentation-related properties that are assessed independently, such as sentiment. Argument reasoning, in contrast, strongly relies on basic inference rules and general world knowledge. Finally, the knowledge used in argument generation seems to be highly task- and domain-dependent.

Not only the types of knowledge but also the techniques employed for injecting that knowledge into CA models substantially differ across the subareas. Considering linguistic knowledge, for example, argument assessment approaches predominantly use lexical cues and traditional symbolic text representations, whereas the body of work on argument reasoning primarily relies on latent semantic representations (i.e., embeddings).    
Most variation in terms of knowledge modeling techniques is found in the argument generation area. Here, the techniques range from template- and structure-based approaches to external lexica and classifiers to embeddings and infusion. %

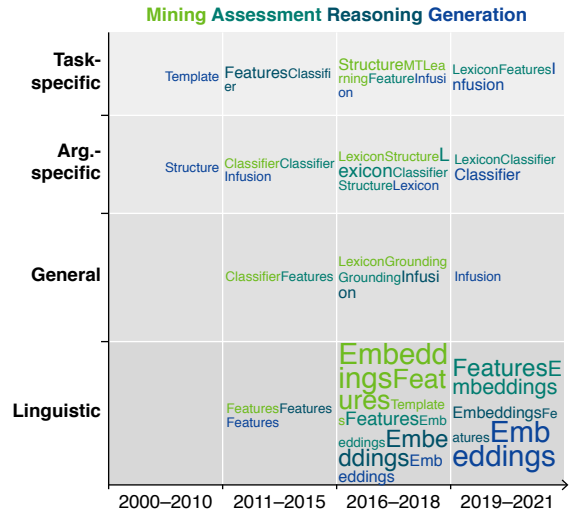
\bsfigure{subtypes}{Techniques of employing knowledge in CA organized by defined time periods (x-axis), knowledge category (y-axis), and area (color). The size of the term indicates the number of occurrences of the techniques (between 1 and 7) in our sample of 40 papers.}

\paragraph{Diachronic Analysis.}

Figure~\ref{subtypes} depicts the temporal development of knowledge modeling techniques in CA, with year, CA subarea, and knowledge type as dimensions. We analyze four time periods, corresponding to pioneering work (2000--2010), the rise of CA in NLP (2011--2015), the shift to distributional methods (2016--2018), and the most recent trends (2019--2021). 

This diachronic analysis reveals that CA is roughly aligned with trends observed in other NLP areas: in the pre-neural era before 2016, knowledge has traditionally been modeled via features, sometimes using knowledge from external resources and outputs or previously trained classifiers (i.e., the pipelined approaches). Later, more advanced techniques such as grounding, infusion, and above all embeddings became more popular. However, we note that distinct techniques are used for the different knowledge types; embeddings, in particular, have been used exclusively to encode linguistic knowledge. Although representation learning can be applied to other argumentative resources, CA efforts in this direction have been few and far between \citep[e.g.,][]{toledo-ronen-etal-2016-expert,alkhatib:2020a}. This warrants more CA work on embedding structured knowledge and towards a unified argumentative representation space that would support the whole spectrum of CA tasks.

\section{Where Should We Go from Here?}
\label{sec:future}
Mastering argumentative discourse requires various types of advanced knowledge~\citep{moens2018argumentation}, making CA one of the most complex problems in AI \citep{atkinson2017towards}. This raises the question of a suitable path to reaching argumentative proficiency for computational models. In this survey, we identified empirical evidence that integrating advanced knowledge can lead to performance improvements on a range of CA tasks. In the following, we pick out those that we see as key ideas toward the goal of mastering argumentation computationally.

Argument mining is often seen as a structure-oriented task. \newcite{lawrence-reed-2017-mining} brought up the notion that topic knowledge may actually predict relations between argument components. \citet{eger-etal-2017-neural}, on the other hand, formulated mining of argument structure as an end-to-end task. Integrating these two views and combining respective methods could hold much promise. 

Despite an abundance of work on encoding and leveraging common sense knowledge~\citep[e.g.,][]{lauscher-etal-2020-common,lin2021common}, argument assessment methods fail to decompose arguments into concepts, with the work of \citet{bar-haim-etal-2017-stance} on stance classification as the positive exception. Despite some evidence of difficulty of integration of common-sense knowledge in  argument reasoning tasks \cite{botschen-etal-2018-frame}, there is no alternative to accurately representing/encoding common-sense knowledge, if we are to build reliable CA systems.
Beyond that, \citet{kobbe-etal-2020-exploring} looked at the impact of morals on argument quality. Such research on modeling fine-grained and socially and culturally-dependent knowledge, such as values and social norms---across languages, is still in its infancy in NLP in general. Systematic research on building respective knowledge sources and benchmarks could push CA to the next level. 

As emphasized by existing work~\citep[e.g.,][]{stede2018argumentation}, argumentation is inherently social and thus highly dependent on the relationship between the speaker and her audience. A more straightforward integration of knowledge about the speaker could prove beneficial: the work of \citet{alshomary-etal-2021-belief}, encoding speaker's belief in argument generation, is a step in this direction. 

\gog{In sum, what we believe is missing in existing work and what could drive the future of CA is a \textit{unified knowledge representation space} that would aggregate and consolidate all CA-relevant knowledge, and be universally beneficial across CA tasks. As shown in this survey, CA-relevant knowledge is fragmented across heterogeneous sources (e.g., corpora, knowledge bases, lexicons) and coupled only sporadically and in an ad-hoc (not principled) manner. Considering the modest sizes of existing CA resources, a methodological orientation to modular and sample-efficient learning and adaptation \cite{houlsby2019parameter,gururangan2020don,ponti2022combining} could provide means to this end.}

\section{Conclusion}
\label{sec:conclusion}
Motivated by the theoretical importance of knowledge in argumentation and by previous work pointing to the need for more research on incorporating advanced types of knowledge in computational argumentation, we have studied the role of knowledge in the body of research works in the field. In total, we surveyed 162 publications spanning the subareas of argument mining, assessment, reasoning, and generation. To organize the approaches described in these works, we proposed a pyramid-like knowledge taxonomy systematizing the types of knowledge according to their specificity, from basic linguistic to task-specific knowledge.

Our survey yields important findings. Many approaches employing advanced knowledge types (e.g., world and argumentation-specific knowledge) report empirical gains. Still, reliance on such external knowledge types is far from uniform across CA areas: While exploitation of such knowledge is pervasive in argument reasoning and generation, it is far less present in argument mining. 
We hope that our findings lead to more systematic consideration of different knowledge sources for CA tasks.

\bibliography{tacl2021}

\begin{thebibliography}{220}
\expandafter\ifx\csname natexlab\endcsname\relax\def\natexlab#1{#1}\fi

\bibitem[{Accuosto and Saggion(2019)}]{accuosto-saggion-2019-transferring}
Pablo Accuosto and Horacio Saggion. 2019.
\newblock \href {https://doi.org/10.18653/v1/W19-4505} {Transferring knowledge
  from discourse to arguments: A case study with scientific abstracts}.
\newblock In \emph{Proceedings of the 6th Workshop on Argument Mining}, pages
  41--51, Florence, Italy. Association for Computational Linguistics.

\bibitem[{Ajjour et~al.(2019)Ajjour, Alshomary, Wachsmuth, and
  Stein}]{ajjour-etal-2019-modeling}
Yamen Ajjour, Milad Alshomary, Henning Wachsmuth, and Benno Stein. 2019.
\newblock \href {https://doi.org/10.18653/v1/D19-1290} {Modeling frames in
  argumentation}.
\newblock In \emph{Proceedings of the 2019 Conference on Empirical Methods in
  Natural Language Processing and the 9th International Joint Conference on
  Natural Language Processing (EMNLP-IJCNLP)}, pages 2922--2932, Hong Kong,
  China. Association for Computational Linguistics.

\bibitem[{Ajjour et~al.(2017)Ajjour, Chen, Kiesel, Wachsmuth, and
  Stein}]{ajjour-etal-2017-unit}
Yamen Ajjour, Wei-Fan Chen, Johannes Kiesel, Henning Wachsmuth, and Benno
  Stein. 2017.
\newblock \href {https://doi.org/10.18653/v1/W17-5115} {Unit segmentation of
  argumentative texts}.
\newblock In \emph{Proceedings of the 4th Workshop on Argument Mining}, pages
  118--128, Copenhagen, Denmark. Association for Computational Linguistics.

\bibitem[{Aker et~al.(2017)Aker, Sliwa, Ma, Lui, Borad, Ziyaei, and
  Ghobadi}]{aker-etal-2017-works}
Ahmet Aker, Alfred Sliwa, Yuan Ma, Ruishen Lui, Niravkumar Borad, Seyedeh
  Ziyaei, and Mina Ghobadi. 2017.
\newblock \href {https://doi.org/10.18653/v1/W17-5112} {What works and what
  does not: Classifier and feature analysis for argument mining}.
\newblock In \emph{Proceedings of the 4th Workshop on Argument Mining}, pages
  91--96, Copenhagen, Denmark. Association for Computational Linguistics.

\bibitem[{Al~Khatib et~al.(2021)Al~Khatib, Ghosal, Hou, de~Waard, and
  Freitag}]{al-khatib-etal-2021-argument}
Khalid Al~Khatib, Tirthankar Ghosal, Yufang Hou, Anita de~Waard, and Dayne
  Freitag. 2021.
\newblock \href {https://doi.org/10.18653/v1/2021.sdp-1.7} {Argument mining for
  scholarly document processing: Taking stock and looking ahead}.
\newblock In \emph{Proceedings of the Second Workshop on Scholarly Document
  Processing}, pages 56--65, Online. Association for Computational Linguistics.

\bibitem[{Al~Khatib et~al.(2020{\natexlab{a}})Al~Khatib, Hou, Wachsmuth,
  Jochim, Bonin, and Stein}]{alkhatib:2020a}
Khalid Al~Khatib, Yufang Hou, Henning Wachsmuth, Charles Jochim, Francesca
  Bonin, and Benno Stein. 2020{\natexlab{a}}.
\newblock End-to-end argumentation knowledge graph construction.
\newblock In \emph{Proceedings of the Thirty-Fourth AAAI Conference on
  Artificial Intelligence}, pages 7367--7374. AAAI.

\bibitem[{Al~Khatib et~al.(2020{\natexlab{b}})Al~Khatib, V{\"o}lske, Syed,
  Kolyada, and Stein}]{al-khatib-etal-2020-exploiting}
Khalid Al~Khatib, Michael V{\"o}lske, Shahbaz Syed, Nikolay Kolyada, and Benno
  Stein. 2020{\natexlab{b}}.
\newblock \href {https://doi.org/10.18653/v1/2020.acl-main.632} {Exploiting
  personal characteristics of debaters for predicting persuasiveness}.
\newblock In \emph{Proceedings of the 58th Annual Meeting of the Association
  for Computational Linguistics}, pages 7067--7072, Online. Association for
  Computational Linguistics.

\bibitem[{Al~Khatib et~al.(2016)Al~Khatib, Wachsmuth, Hagen, K{\"o}hler, and
  Stein}]{al-khatib-etal-2016-cross}
Khalid Al~Khatib, Henning Wachsmuth, Matthias Hagen, Jonas K{\"o}hler, and
  Benno Stein. 2016.
\newblock \href {https://doi.org/10.18653/v1/N16-1165} {Cross-domain mining of
  argumentative text through distant supervision}.
\newblock In \emph{Proceedings of the 2016 Conference of the North {A}merican
  Chapter of the Association for Computational Linguistics: Human Language
  Technologies}, pages 1395--1404, San Diego, California. Association for
  Computational Linguistics.

\bibitem[{Alshomary et~al.(2021)Alshomary, Chen, Gurcke, and
  Wachsmuth}]{alshomary-etal-2021-belief}
Milad Alshomary, Wei-Fan Chen, Timon Gurcke, and Henning Wachsmuth. 2021.
\newblock \href {https://www.aclweb.org/anthology/2021.eacl-main.17}
  {Belief-based generation of argumentative claims}.
\newblock In \emph{Proceedings of the 16th Conference of the European Chapter
  of the Association for Computational Linguistics: Main Volume}, pages
  224--233, Online. Association for Computational Linguistics.

\bibitem[{Alshomary et~al.(2020{\natexlab{a}})Alshomary, D\"{u}sterhus, and
  Wachsmuth}]{10.1145/3397271.3401186}
Milad Alshomary, Nick D\"{u}sterhus, and Henning Wachsmuth. 2020{\natexlab{a}}.
\newblock \href {https://doi.org/10.1145/3397271.3401186} {Extractive snippet
  generation for arguments}.
\newblock In \emph{Proceedings of the 43rd International ACM SIGIR Conference
  on Research and Development in Information Retrieval}, SIGIR '20, page
  1969–1972, New York, NY, USA. Association for Computing Machinery.

\bibitem[{Alshomary et~al.(2020{\natexlab{b}})Alshomary, Syed, Potthast, and
  Wachsmuth}]{alshomary-etal-2020-target}
Milad Alshomary, Shahbaz Syed, Martin Potthast, and Henning Wachsmuth.
  2020{\natexlab{b}}.
\newblock \href {https://doi.org/10.18653/v1/2020.acl-main.399} {Target
  inference in argument conclusion generation}.
\newblock In \emph{Proceedings of the 58th Annual Meeting of the Association
  for Computational Linguistics}, pages 4334--4345, Online. Association for
  Computational Linguistics.

\bibitem[{Aristotle(ca. 350 \textsc{B.C.E.}/ translated
  2007)}]{kennedy2007aristotle}
Aristotle. ca. 350 \textsc{B.C.E.}/ translated 2007.
\newblock \emph{On Rhetoric: A Theory of Civic Discourse}.
\newblock Oxford University Press, Oxford, UK.
\newblock Translated by George A. Kennedy.

\bibitem[{Athar(2011)}]{athar_sentiment_2011}
Awais Athar. 2011.
\newblock Sentiment {{Analysis}} of {{Citations Using Sentence
  Structure}}-based {{Features}}.
\newblock In \emph{Proceedings of the {{ACL}} 2011 {{Student Session}}}, HLT-SS
  '11, pages 81--87, Stroudsburg, PA, USA. {Association for Computational
  Linguistics}.

\bibitem[{Atkinson et~al.(2017)Atkinson, Baroni, Giacomin, Hunter, Prakken,
  Reed, Simari, Thimm, and Villata}]{atkinson2017towards}
Katie Atkinson, Pietro Baroni, Massimiliano Giacomin, Anthony Hunter, Henry
  Prakken, Chris Reed, Guillermo Simari, Matthias Thimm, and Serena Villata.
  2017.
\newblock \href {https://doi.org/10.1609/aimag.v38i3.2704} {Towards artificial
  argumentation}.
\newblock \emph{AI Magazine}, 38(3):25--36.

\bibitem[{Bar-Haim et~al.(2017{\natexlab{a}})Bar-Haim, Bhattacharya, Dinuzzo,
  Saha, and Slonim}]{bar-haim-etal-2017-stance}
Roy Bar-Haim, Indrajit Bhattacharya, Francesco Dinuzzo, Amrita Saha, and Noam
  Slonim. 2017{\natexlab{a}}.
\newblock \href {https://www.aclweb.org/anthology/E17-1024} {Stance
  classification of context-dependent claims}.
\newblock In \emph{Proceedings of the 15th Conference of the {E}uropean Chapter
  of the Association for Computational Linguistics: Volume 1, Long Papers},
  pages 251--261, Valencia, Spain. Association for Computational Linguistics.

\bibitem[{Bar-Haim et~al.(2017{\natexlab{b}})Bar-Haim, Edelstein, Jochim, and
  Slonim}]{bar-haim-etal-2017-improving}
Roy Bar-Haim, Lilach Edelstein, Charles Jochim, and Noam Slonim.
  2017{\natexlab{b}}.
\newblock \href {https://doi.org/10.18653/v1/W17-5104} {Improving claim stance
  classification with lexical knowledge expansion and context utilization}.
\newblock In \emph{Proceedings of the 4th Workshop on Argument Mining}, pages
  32--38, Copenhagen, Denmark. Association for Computational Linguistics.

\bibitem[{Bar-Haim et~al.(2020)Bar-Haim, Kantor, Eden, Friedman, Lahav, and
  Slonim}]{bar-haim-etal-2020-quantitative}
Roy Bar-Haim, Yoav Kantor, Lilach Eden, Roni Friedman, Dan Lahav, and Noam
  Slonim. 2020.
\newblock \href {https://doi.org/10.18653/v1/2020.emnlp-main.3} {Quantitative
  argument summarization and beyond: Cross-domain key point analysis}.
\newblock In \emph{Proceedings of the 2020 Conference on Empirical Methods in
  Natural Language Processing (EMNLP)}, pages 39--49, Online. Association for
  Computational Linguistics.

\bibitem[{Becker et~al.(2021)Becker, Liang, and
  Frank}]{becker-etal-2021-reconstructing}
Maria Becker, Siting Liang, and Anette Frank. 2021.
\newblock \href {https://doi.org/10.18653/v1/2021.deelio-1.2} {Reconstructing
  implicit knowledge with language models}.
\newblock In \emph{Proceedings of Deep Learning Inside Out (DeeLIO): The 2nd
  Workshop on Knowledge Extraction and Integration for Deep Learning
  Architectures}, pages 11--24, Online. Association for Computational
  Linguistics.

\bibitem[{Bentahar et~al.(2010{\natexlab{a}})Bentahar, Moulin, and
  B{\'e}langer}]{bentahar2010taxonomy}
Jamal Bentahar, Bernard Moulin, and Micheline B{\'e}langer. 2010{\natexlab{a}}.
\newblock A taxonomy of argumentation models used for knowledge representation.
\newblock \emph{Artificial Intelligence Review}, 33(3):211--259.

\bibitem[{Bentahar et~al.(2010{\natexlab{b}})Bentahar, Moulin, and
  Bélanger}]{bentaharTaxonomyArgumentationModels2010}
Jamal Bentahar, Bernard Moulin, and Micheline Bélanger. 2010{\natexlab{b}}.
\newblock \href {https://doi.org/10.1007/s10462-010-9154-1} {A taxonomy of
  argumentation models used for knowledge representation}.
\newblock \emph{Artificial Intelligence Review}, 33(3):211--259.

\bibitem[{Bilu et~al.(2019)Bilu, Gera, Hershcovich, Sznajder, Lahav,
  Moshkowich, Malet, Gavron, and Slonim}]{bilu-etal-2019-argument}
Yonatan Bilu, Ariel Gera, Daniel Hershcovich, Benjamin Sznajder, Dan Lahav, Guy
  Moshkowich, Anael Malet, Assaf Gavron, and Noam Slonim. 2019.
\newblock \href {https://doi.org/10.18653/v1/P19-1097} {Argument invention from
  first principles}.
\newblock In \emph{Proceedings of the 57th Annual Meeting of the Association
  for Computational Linguistics}, pages 1013--1026, Florence, Italy.
  Association for Computational Linguistics.

\bibitem[{Bilu and Slonim(2016)}]{bilu-slonim-2016-claim}
Yonatan Bilu and Noam Slonim. 2016.
\newblock \href {https://doi.org/10.18653/v1/P16-2085} {Claim synthesis via
  predicate recycling}.
\newblock In \emph{Proceedings of the 54th Annual Meeting of the Association
  for Computational Linguistics (Volume 2: Short Papers)}, pages 525--530,
  Berlin, Germany. Association for Computational Linguistics.

\bibitem[{Bojanowski et~al.(2017)Bojanowski, Grave, Joulin, and
  Mikolov}]{bojanowski2017enriching}
Piotr Bojanowski, Edouard Grave, Armand Joulin, and Tomas Mikolov. 2017.
\newblock Enriching word vectors with subword information.
\newblock \emph{Transactions of the Association for Computational Linguistics},
  5:135--146.

\bibitem[{Boltu{\v{z}}i{\'c} and
  {\v{S}}najder(2014)}]{boltuzic-snajder-2014-back}
Filip Boltu{\v{z}}i{\'c} and Jan {\v{S}}najder. 2014.
\newblock \href {https://doi.org/10.3115/v1/W14-2107} {Back up your stance:
  {{R}}ecognizing arguments in online discussions}.
\newblock In \emph{Proceedings of the First Workshop on Argumentation Mining},
  pages 49--58, Baltimore, Maryland. Association for Computational Linguistics.

\bibitem[{Boltu{\v{z}}i{\'c} and
  {\v{S}}najder(2016)}]{boltuzic-snajder-2016-fill}
Filip Boltu{\v{z}}i{\'c} and Jan {\v{S}}najder. 2016.
\newblock \href {https://doi.org/10.18653/v1/W16-2815} {Fill the gap!
  {{A}}nalyzing implicit premises between claims from online debates}.
\newblock In \emph{Proceedings of the Third Workshop on Argument Mining
  ({A}rg{M}ining2016)}, pages 124--133, Berlin, Germany. Association for
  Computational Linguistics.

\bibitem[{Boltu{\v{z}}i{\'c} and
  {\v{S}}najder(2017)}]{boltuzic-snajder-2017-toward}
Filip Boltu{\v{z}}i{\'c} and Jan {\v{S}}najder. 2017.
\newblock \href {https://doi.org/10.18653/v1/W17-5210} {Toward stance
  classification based on claim microstructures}.
\newblock In \emph{Proceedings of the 8th Workshop on Computational Approaches
  to Subjectivity, Sentiment and Social Media Analysis}, pages 74--80,
  Copenhagen, Denmark. Association for Computational Linguistics.

\bibitem[{Botschen et~al.(2018)Botschen, Sorokin, and
  Gurevych}]{botschen-etal-2018-frame}
Teresa Botschen, Daniil Sorokin, and Iryna Gurevych. 2018.
\newblock \href {https://doi.org/10.18653/v1/W18-5211} {Frame- and entity-based
  knowledge for common-sense argumentative reasoning}.
\newblock In \emph{Proceedings of the 5th Workshop on Argument Mining}, pages
  90--96, Brussels, Belgium. Association for Computational Linguistics.

\bibitem[{Bowman et~al.(2015)Bowman, Angeli, Potts, and
  Manning}]{bowman2015large}
Samuel Bowman, Gabor Angeli, Christopher Potts, and Christopher~D Manning.
  2015.
\newblock A large annotated corpus for learning natural language inference.
\newblock In \emph{Proceedings of the 2015 Conference on Empirical Methods in
  Natural Language Processing}, pages 632--642.

\bibitem[{Brassard et~al.(2018)Brassard, Kuculo, Boltu{\v{z}}i{\'c}, and
  {\v{S}}najder}]{brassard-etal-2018-takelab}
Ana Brassard, Tin Kuculo, Filip Boltu{\v{z}}i{\'c}, and Jan {\v{S}}najder.
  2018.
\newblock \href {https://doi.org/10.18653/v1/S18-1192} {{T}ake{L}ab at
  {S}em{E}val-2018 task12: Argument reasoning comprehension with skip-thought
  vectors}.
\newblock In \emph{Proceedings of The 12th International Workshop on Semantic
  Evaluation}, pages 1133--1136, New Orleans, Louisiana. Association for
  Computational Linguistics.

\bibitem[{Cabrio and Villata(2012)}]{cabrio-villata-2012-combining}
Elena Cabrio and Serena Villata. 2012.
\newblock \href {https://www.aclweb.org/anthology/P12-2041} {Combining textual
  entailment and argumentation theory for supporting online debates
  interactions}.
\newblock In \emph{Proceedings of the 50th Annual Meeting of the Association
  for Computational Linguistics (Volume 2: Short Papers)}, pages 208--212, Jeju
  Island, Korea. Association for Computational Linguistics.

\bibitem[{Cabrio and Villata(2018)}]{cabrio2018five}
Elena Cabrio and Serena Villata. 2018.
\newblock Five years of argument mining: a data-driven analysis.
\newblock In \emph{IJCAI}, volume~18, pages 5427--5433.

\bibitem[{Carenini and Moore(2006)}]{CARENINI2006925}
Giuseppe Carenini and Johanna~D. Moore. 2006.
\newblock \href {https://doi.org/https://doi.org/10.1016/j.artint.2006.05.003}
  {Generating and evaluating evaluative arguments}.
\newblock \emph{Artificial Intelligence}, 170(11):925--952.

\bibitem[{Carstens and Toni(2015)}]{carstens-toni-2015-towards}
Lucas Carstens and Francesca Toni. 2015.
\newblock \href {https://doi.org/10.3115/v1/W15-0504} {Towards relation based
  argumentation mining}.
\newblock In \emph{Proceedings of the 2nd Workshop on Argumentation Mining},
  pages 29--34, Denver, CO. Association for Computational Linguistics.

\bibitem[{Chakrabarty et~al.(2019)Chakrabarty, Hidey, Muresan, McKeown, and
  Hwang}]{chakrabarty-etal-2019-ampersand}
Tuhin Chakrabarty, Christopher Hidey, Smaranda Muresan, Kathy McKeown, and
  Alyssa Hwang. 2019.
\newblock \href {https://doi.org/10.18653/v1/D19-1291} {{AMPERSAND}: Argument
  mining for {PERS}u{A}sive o{N}line discussions}.
\newblock In \emph{Proceedings of the 2019 Conference on Empirical Methods in
  Natural Language Processing and the 9th International Joint Conference on
  Natural Language Processing (EMNLP-IJCNLP)}, pages 2933--2943, Hong Kong,
  China. Association for Computational Linguistics.

\bibitem[{Chalaguine and Schulz(2017)}]{chalaguine-schulz-2017-assessing}
Lisa~Andreevna Chalaguine and Claudia Schulz. 2017.
\newblock \href {https://www.aclweb.org/anthology/E17-4008} {Assessing
  convincingness of arguments in online debates with limited number of
  features}.
\newblock In \emph{Proceedings of the Student Research Workshop at the 15th
  Conference of the {E}uropean Chapter of the Association for Computational
  Linguistics}, pages 75--83, Valencia, Spain. Association for Computational
  Linguistics.

\bibitem[{Chen et~al.(2018)Chen, Wachsmuth, Al-Khatib, and
  Stein}]{chen-etal-2018-learning}
Wei-Fan Chen, Henning Wachsmuth, Khalid Al-Khatib, and Benno Stein. 2018.
\newblock \href {https://doi.org/10.18653/v1/W18-6509} {Learning to flip the
  bias of news headlines}.
\newblock In \emph{Proceedings of the 11th International Conference on Natural
  Language Generation}, pages 79--88, Tilburg University, The Netherlands.
  Association for Computational Linguistics.

\bibitem[{Choi and Lee(2018)}]{choi-lee-2018-gist}
HongSeok Choi and Hyunju Lee. 2018.
\newblock \href {https://doi.org/10.18653/v1/S18-1122} {{GIST} at
  {S}em{E}val-2018 task 12: A network transferring inference knowledge to
  argument reasoning comprehension task}.
\newblock In \emph{Proceedings of The 12th International Workshop on Semantic
  Evaluation}, pages 773--777, New Orleans, Louisiana. Association for
  Computational Linguistics.

\bibitem[{Clark et~al.(2020)Clark, Luong, Le, and Manning}]{clark2020electra}
Kevin Clark, Minh-Thang Luong, Quoc~V Le, and Christopher~D Manning. 2020.
\newblock Electra: Pre-training text encoders as discriminators rather than
  generators.
\newblock In \emph{International Conference on Learning Representations}.

\bibitem[{Cocarascu and Toni(2017)}]{cocarascu-toni-2017-identifying}
Oana Cocarascu and Francesca Toni. 2017.
\newblock \href {https://doi.org/10.18653/v1/D17-1144} {Identifying attack and
  support argumentative relations using deep learning}.
\newblock In \emph{Proceedings of the 2017 Conference on Empirical Methods in
  Natural Language Processing}, pages 1374--1379, Copenhagen, Denmark.
  Association for Computational Linguistics.

\bibitem[{Dagan et~al.(2013)Dagan, Roth, Sammons, and
  Zanzotto}]{dagan2013recognizing}
Ido Dagan, Dan Roth, Mark Sammons, and Fabio~Massimo Zanzotto. 2013.
\newblock Recognizing textual entailment: Models and applications.
\newblock \emph{Synthesis Lectures on Human Language Technologies},
  6(4):1--220.

\bibitem[{Daxenberger et~al.(2017)Daxenberger, Eger, Habernal, Stab, and
  Gurevych}]{daxenberger-etal-2017-essence}
Johannes Daxenberger, Steffen Eger, Ivan Habernal, Christian Stab, and Iryna
  Gurevych. 2017.
\newblock \href {https://doi.org/10.18653/v1/D17-1218} {What is the essence of
  a claim? {{C}}ross-domain claim identification}.
\newblock In \emph{Proceedings of the 2017 Conference on Empirical Methods in
  Natural Language Processing}, pages 2055--2066, Copenhagen, Denmark.
  Association for Computational Linguistics.

\bibitem[{Delobelle et~al.(2019)Delobelle, Cunha, Massip~Cano, Peperkamp, and
  Berendt}]{delobelle-etal-2019-computational}
Pieter Delobelle, Murilo Cunha, Eric Massip~Cano, Jeroen Peperkamp, and Bettina
  Berendt. 2019.
\newblock \href {https://doi.org/10.18653/v1/P19-2028} {Computational ad
  hominem detection}.
\newblock In \emph{Proceedings of the 57th Annual Meeting of the Association
  for Computational Linguistics: Student Research Workshop}, pages 203--209,
  Florence, Italy. Association for Computational Linguistics.

\bibitem[{Devlin et~al.(2019)Devlin, Chang, Lee, and
  Toutanova}]{devlin-etal-2019-bert}
Jacob Devlin, Ming-Wei Chang, Kenton Lee, and Kristina Toutanova. 2019.
\newblock \href {https://doi.org/10.18653/v1/N19-1423} {{BERT}: Pre-training of
  deep bidirectional transformers for language understanding}.
\newblock In \emph{Proceedings of the 2019 Conference of the North {A}merican
  Chapter of the Association for Computational Linguistics: Human Language
  Technologies, Volume 1 (Long and Short Papers)}, pages 4171--4186,
  Minneapolis, Minnesota. Association for Computational Linguistics.

\bibitem[{Dretske(1981)}]{Dretske1981-DREKAT}
Fred~I. Dretske. 1981.
\newblock \emph{Knowledge and the Flow of Information}.
\newblock MIT Press.

\bibitem[{Dumani and Schenkel(2019)}]{10.1145/3331184.3331282}
Lorik Dumani and Ralf Schenkel. 2019.
\newblock \href {https://doi.org/10.1145/3331184.3331282} {A systematic
  comparison of methods for finding good premises for claims}.
\newblock In \emph{Proceedings of the 42nd International ACM SIGIR Conference
  on Research and Development in Information Retrieval}, SIGIR'19, page
  957–960, New York, NY, USA. Association for Computing Machinery.

\bibitem[{Dung(1995)}]{dung1995acceptability}
Phan~Minh Dung. 1995.
\newblock On the acceptability of arguments and its fundamental role in
  nonmonotonic reasoning, logic programming and n-person games.
\newblock \emph{Artificial intelligence}, 77(2):321--357.

\bibitem[{Durmus and Cardie(2018)}]{durmus-cardie-2018-exploring}
Esin Durmus and Claire Cardie. 2018.
\newblock \href {https://doi.org/10.18653/v1/N18-1094} {Exploring the role of
  prior beliefs for argument persuasion}.
\newblock In \emph{Proceedings of the 2018 Conference of the North {A}merican
  Chapter of the Association for Computational Linguistics: Human Language
  Technologies, Volume 1 (Long Papers)}, pages 1035--1045, New Orleans,
  Louisiana. Association for Computational Linguistics.

\bibitem[{Durmus and Cardie(2019)}]{durmus-cardie-2019-corpus}
Esin Durmus and Claire Cardie. 2019.
\newblock \href {https://doi.org/10.18653/v1/P19-1057} {A corpus for modeling
  user and language effects in argumentation on online debating}.
\newblock In \emph{Proceedings of the 57th Annual Meeting of the Association
  for Computational Linguistics}, pages 602--607, Florence, Italy. Association
  for Computational Linguistics.

\bibitem[{Durmus et~al.(2019)Durmus, Ladhak, and
  Cardie}]{durmus-etal-2019-determining}
Esin Durmus, Faisal Ladhak, and Claire Cardie. 2019.
\newblock \href {https://doi.org/10.18653/v1/P19-1456} {Determining relative
  argument specificity and stance for complex argumentative structures}.
\newblock In \emph{Proceedings of the 57th Annual Meeting of the Association
  for Computational Linguistics}, pages 4630--4641, Florence, Italy.
  Association for Computational Linguistics.

\bibitem[{Dusmanu et~al.(2017)Dusmanu, Cabrio, and
  Villata}]{dusmanu-etal-2017-argument}
Mihai Dusmanu, Elena Cabrio, and Serena Villata. 2017.
\newblock \href {https://doi.org/10.18653/v1/D17-1245} {Argument mining on
  {T}witter: Arguments, facts and sources}.
\newblock In \emph{Proceedings of the 2017 Conference on Empirical Methods in
  Natural Language Processing}, pages 2317--2322, Copenhagen, Denmark.
  Association for Computational Linguistics.

\bibitem[{Egan et~al.(2016)Egan, Siddharthan, and
  Wyner}]{egan-etal-2016-summarising}
Charlie Egan, Advaith Siddharthan, and Adam Wyner. 2016.
\newblock \href {https://doi.org/10.18653/v1/W16-2816} {Summarising the points
  made in online political debates}.
\newblock In \emph{Proceedings of the Third Workshop on Argument Mining
  ({A}rg{M}ining2016)}, pages 134--143, Berlin, Germany. Association for
  Computational Linguistics.

\bibitem[{Eger et~al.(2017)Eger, Daxenberger, and
  Gurevych}]{eger-etal-2017-neural}
Steffen Eger, Johannes Daxenberger, and Iryna Gurevych. 2017.
\newblock \href {https://doi.org/10.18653/v1/P17-1002} {Neural end-to-end
  learning for computational argumentation mining}.
\newblock In \emph{Proceedings of the 55th Annual Meeting of the Association
  for Computational Linguistics (Volume 1: Long Papers)}, pages 11--22,
  Vancouver, Canada. Association for Computational Linguistics.

\bibitem[{Eger et~al.(2018)Eger, Daxenberger, Stab, and
  Gurevych}]{eger-etal-2018-cross}
Steffen Eger, Johannes Daxenberger, Christian Stab, and Iryna Gurevych. 2018.
\newblock \href {https://www.aclweb.org/anthology/C18-1071} {Cross-lingual
  argumentation mining: Machine translation (and a bit of projection) is all
  you need!}
\newblock In \emph{Proceedings of the 27th International Conference on
  Computational Linguistics}, pages 831--844, Santa Fe, New Mexico, USA.
  Association for Computational Linguistics.

\bibitem[{Eide(2019)}]{eide-2019-swedish}
Stian~R{\o}dven Eide. 2019.
\newblock \href {https://doi.org/10.18653/v1/W19-4506} {The {S}wedish
  {P}oli{G}raph: A semantic graph for argument mining of {S}wedish
  parliamentary data}.
\newblock In \emph{Proceedings of the 6th Workshop on Argument Mining}, pages
  52--57, Florence, Italy. Association for Computational Linguistics.

\bibitem[{El~Baff et~al.(2019)El~Baff, Wachsmuth, Al~Khatib, Stede, and
  Stein}]{el-baff-etal-2019-computational}
Roxanne El~Baff, Henning Wachsmuth, Khalid Al~Khatib, Manfred Stede, and Benno
  Stein. 2019.
\newblock \href {https://doi.org/10.18653/v1/W19-8607} {Computational
  argumentation synthesis as a language modeling task}.
\newblock In \emph{Proceedings of the 12th International Conference on Natural
  Language Generation}, pages 54--64, Tokyo, Japan. Association for
  Computational Linguistics.

\bibitem[{El~Baff et~al.(2018)El~Baff, Wachsmuth, Al-Khatib, and
  Stein}]{el-baff-etal-2018-challenge}
Roxanne El~Baff, Henning Wachsmuth, Khalid Al-Khatib, and Benno Stein. 2018.
\newblock \href {https://doi.org/10.18653/v1/K18-1044} {Challenge or empower:
  Revisiting argumentation quality in a news editorial corpus}.
\newblock In \emph{Proceedings of the 22nd Conference on Computational Natural
  Language Learning}, pages 454--464, Brussels, Belgium. Association for
  Computational Linguistics.

\bibitem[{El~Baff et~al.(2020)El~Baff, Wachsmuth, Al~Khatib, and
  Stein}]{el-baff-etal-2020-analyzing}
Roxanne El~Baff, Henning Wachsmuth, Khalid Al~Khatib, and Benno Stein. 2020.
\newblock \href {https://doi.org/10.18653/v1/2020.acl-main.287} {{A}nalyzing
  the {P}ersuasive {E}ffect of {S}tyle in {N}ews {E}ditorial {A}rgumentation}.
\newblock In \emph{Proceedings of the 58th Annual Meeting of the Association
  for Computational Linguistics}, pages 3154--3160, Online. Association for
  Computational Linguistics.

\bibitem[{Feng and Hirst(2011)}]{feng-hirst-2011-classifying}
Vanessa~Wei Feng and Graeme Hirst. 2011.
\newblock \href {https://www.aclweb.org/anthology/P11-1099} {Classifying
  arguments by scheme}.
\newblock In \emph{Proceedings of the 49th Annual Meeting of the Association
  for Computational Linguistics: Human Language Technologies}, pages 987--996,
  Portland, Oregon, USA. Association for Computational Linguistics.

\bibitem[{Forbes et~al.(2020)Forbes, Hwang, Shwartz, Sap, and
  Choi}]{forbes-etal-2020-social}
Maxwell Forbes, Jena~D. Hwang, Vered Shwartz, Maarten Sap, and Yejin Choi.
  2020.
\newblock \href {https://doi.org/10.18653/v1/2020.emnlp-main.48} {Social
  chemistry 101: Learning to reason about social and moral norms}.
\newblock In \emph{Proceedings of the 2020 Conference on Empirical Methods in
  Natural Language Processing (EMNLP)}, pages 653--670, Online. Association for
  Computational Linguistics.

\bibitem[{Freeman(2011)}]{freeman:2011}
James~B. Freeman. 2011.
\newblock \emph{Argument Structure: {R}epresentation and Theory}.
\newblock Springer.

\bibitem[{Galassi et~al.(2018)Galassi, Lippi, and
  Torroni}]{galassi-etal-2018-argumentative}
Andrea Galassi, Marco Lippi, and Paolo Torroni. 2018.
\newblock \href {https://doi.org/10.18653/v1/W18-5201} {Argumentative link
  prediction using residual networks and multi-objective learning}.
\newblock In \emph{Proceedings of the 5th Workshop on Argument Mining}, pages
  1--10, Brussels, Belgium. Association for Computational Linguistics.

\bibitem[{Gatt and Krahmer(2018)}]{gatt2018survey}
Albert Gatt and Emiel Krahmer. 2018.
\newblock Survey of the state of the art in natural language generation: Core
  tasks, applications and evaluation.
\newblock \emph{Journal of Artificial Intelligence Research}, 61:65--170.

\bibitem[{Gemechu and Reed(2019)}]{gemechu-reed-2019-decompositional}
Debela Gemechu and Chris Reed. 2019.
\newblock \href {https://doi.org/10.18653/v1/P19-1049} {Decompositional
  argument mining: A general purpose approach for argument graph construction}.
\newblock In \emph{Proceedings of the 57th Annual Meeting of the Association
  for Computational Linguistics}, pages 516--526, Florence, Italy. Association
  for Computational Linguistics.

\bibitem[{Ghosh et~al.(2016)Ghosh, Khanam, Han, and
  Muresan}]{ghosh-etal-2016-coarse}
Debanjan Ghosh, Aquila Khanam, Yubo Han, and Smaranda Muresan. 2016.
\newblock \href {https://doi.org/10.18653/v1/P16-2089} {Coarse-grained
  argumentation features for scoring persuasive essays}.
\newblock In \emph{Proceedings of the 54th Annual Meeting of the Association
  for Computational Linguistics (Volume 2: Short Papers)}, pages 549--554,
  Berlin, Germany. Association for Computational Linguistics.

\bibitem[{Gilbert(1977)}]{nigel1977referencing}
G~Nigel Gilbert. 1977.
\newblock Referencing as persuasion.
\newblock \emph{Social Studies of Science}, 7(1):113--122.

\bibitem[{Gleize et~al.(2019)Gleize, Shnarch, Choshen, Dankin, Moshkowich,
  Aharonov, and Slonim}]{gleize-etal-2019-convinced}
Martin Gleize, Eyal Shnarch, Leshem Choshen, Lena Dankin, Guy Moshkowich, Ranit
  Aharonov, and Noam Slonim. 2019.
\newblock \href {https://doi.org/10.18653/v1/P19-1093} {Are you convinced?
  choosing the more convincing evidence with a {S}iamese network}.
\newblock In \emph{Proceedings of the 57th Annual Meeting of the Association
  for Computational Linguistics}, pages 967--976, Florence, Italy. Association
  for Computational Linguistics.

\bibitem[{Goldman(1967)}]{10.2307/2024268}
Alvin~I. Goldman. 1967.
\newblock \href {http://www.jstor.org/stable/2024268} {A causal theory of
  knowing}.
\newblock \emph{The Journal of Philosophy}, 64(12):357--372.

\bibitem[{Gottschalk-Mazouz(2013)}]{gottschalk2013internet}
Niels Gottschalk-Mazouz. 2013.
\newblock Internet and the flow of knowledge: Which ethical and political
  challenges will we face?
\newblock \emph{From ontos verlag: Publications of the Austrian Ludwig
  Wittgenstein Society-New Series (Volumes 1-18)}, 7.

\bibitem[{Graham et~al.(2013)Graham, Haidt, Koleva, Motyl, Iyer, Wojcik, and
  Ditto}]{graham2013moral}
Jesse Graham, Jonathan Haidt, Sena Koleva, Matt Motyl, Ravi Iyer, Sean~P
  Wojcik, and Peter~H Ditto. 2013.
\newblock Moral foundations theory: The pragmatic validity of moral pluralism.
\newblock In \emph{Advances in experimental social psychology}, volume~47,
  pages 55--130. Elsevier.

\bibitem[{Graham et~al.(2009)Graham, Haidt, and Nosek}]{graham2009liberals}
Jesse Graham, Jonathan Haidt, and Brian~A Nosek. 2009.
\newblock Liberals and conservatives rely on different sets of moral
  foundations.
\newblock \emph{Journal of personality and social psychology}, 96(5):1029.

\bibitem[{Gretz et~al.(2020{\natexlab{a}})Gretz, Bilu, Cohen-Karlik, and
  Slonim}]{gretz-etal-2020-workweek}
Shai Gretz, Yonatan Bilu, Edo Cohen-Karlik, and Noam Slonim.
  2020{\natexlab{a}}.
\newblock \href {https://doi.org/10.18653/v1/2020.findings-emnlp.47} {The
  workweek is the best time to start a family {--} a study of {GPT}-2 based
  claim generation}.
\newblock In \emph{Findings of the Association for Computational Linguistics:
  EMNLP 2020}, pages 528--544, Online. Association for Computational
  Linguistics.

\bibitem[{Gretz et~al.(2020{\natexlab{b}})Gretz, Friedman, Cohen-Karlik,
  Toledo, Lahav, Aharonov, and Slonim}]{gretz2020large}
Shai Gretz, Roni Friedman, Edo Cohen-Karlik, Assaf Toledo, Dan Lahav, Ranit
  Aharonov, and Noam Slonim. 2020{\natexlab{b}}.
\newblock A large-scale dataset for argument quality ranking: Construction and
  analysis.
\newblock \emph{Proceedings of the AAAI Conference on Artificial Intelligence},
  34(05):7805--7813.

\bibitem[{Gu et~al.(2018)Gu, Wei, Xu, Fu, Liu, and
  Huang}]{gu-etal-2018-incorporating}
Yunfan Gu, Zhongyu Wei, Maoran Xu, Hao Fu, Yang Liu, and Xuanjing Huang. 2018.
\newblock \href {https://doi.org/10.18653/v1/W18-5212} {Incorporating topic
  aspects for online comment convincingness evaluation}.
\newblock In \emph{Proceedings of the 5th Workshop on Argument Mining}, pages
  97--104, Brussels, Belgium. Association for Computational Linguistics.

\bibitem[{Gururangan et~al.(2020)Gururangan, Marasovi{\'c}, Swayamdipta, Lo,
  Beltagy, Downey, and Smith}]{gururangan2020don}
Suchin Gururangan, Ana Marasovi{\'c}, Swabha Swayamdipta, Kyle Lo, Iz~Beltagy,
  Doug Downey, and Noah~A Smith. 2020.
\newblock Don’t stop pretraining: Adapt language models to domains and tasks.
\newblock In \emph{Proceedings of the 58th Annual Meeting of the Association
  for Computational Linguistics}, pages 8342--8360.

\bibitem[{Habernal et~al.(2014)Habernal, {Eckle-Kohler}, and
  Gurevych}]{habernal2014argumentation}
Ivan Habernal, Judith {Eckle-Kohler}, and Iryna Gurevych. 2014.
\newblock \href {http://ceur-ws.org/Vol-1341/paper4.pdf} {Argumentation
  {{Mining}} on the {{Web}} from {{Information Seeking Perspective}}.}
\newblock In \emph{{{Proceedings of the Workshop on Frontiers and Connections
  between Argumentation Theory and Natural Language Processing}}},
  Forlì-Cesena, Italy.

\bibitem[{Habernal and
  Gurevych(2016{\natexlab{a}})}]{habernal-gurevych-2016-makes}
Ivan Habernal and Iryna Gurevych. 2016{\natexlab{a}}.
\newblock \href {https://doi.org/10.18653/v1/D16-1129} {What makes a convincing
  argument? {{E}}mpirical analysis and detecting attributes of convincingness
  in web argumentation}.
\newblock In \emph{Proceedings of the 2016 Conference on Empirical Methods in
  Natural Language Processing}, pages 1214--1223, Austin, Texas. Association
  for Computational Linguistics.

\bibitem[{Habernal and Gurevych(2016{\natexlab{b}})}]{Habernal2016WhichAI}
Ivan Habernal and Iryna Gurevych. 2016{\natexlab{b}}.
\newblock \href {https://doi.org/10.18653/v1/P16-1150} {Which argument is more
  convincing? {{Analyzing}} and predicting convincingness of web arguments
  using bidirectional {{LSTM}}}.
\newblock In \emph{Proceedings of the 54th Annual Meeting of the Association
  for Computational Linguistics (Volume 1: Long Papers)}, page 11–22, Berlin,
  Germany. Association for Computational Linguistics.

\bibitem[{Habernal and Gurevych(2017)}]{habernal-gurevych-2017-argumentation}
Ivan Habernal and Iryna Gurevych. 2017.
\newblock \href {https://doi.org/10.1162/COLI_a_00276} {Argumentation mining in
  user-generated web discourse}.
\newblock \emph{Computational Linguistics}, 43(1):125--179.

\bibitem[{Habernal et~al.(2018{\natexlab{a}})Habernal, Pauli, and
  Gurevych}]{HABERNAL18.494}
Ivan Habernal, Patrick Pauli, and Iryna Gurevych. 2018{\natexlab{a}}.
\newblock {Adapting Serious Game for Fallacious Argumentation to German:
  Pitfalls, Insights, and Best Practices}.
\newblock In \emph{Proceedings of the Eleventh International Conference on
  Language Resources and Evaluation (LREC 2018)}, Miyazaki, Japan. European
  Language Resources Association (ELRA).

\bibitem[{Habernal et~al.(2018{\natexlab{b}})Habernal, Wachsmuth, Gurevych, and
  Stein}]{habernal-etal-2018-argument}
Ivan Habernal, Henning Wachsmuth, Iryna Gurevych, and Benno Stein.
  2018{\natexlab{b}}.
\newblock \href {https://doi.org/10.18653/v1/N18-1175} {The argument reasoning
  comprehension task: {I}dentification and reconstruction of implicit
  warrants}.
\newblock In \emph{Proceedings of the 2018 Conference of the North {A}merican
  Chapter of the Association for Computational Linguistics: Human Language
  Technologies, Volume 1 (Long Papers)}, pages 1930--1940, New Orleans,
  Louisiana. Association for Computational Linguistics.

\bibitem[{Habernal et~al.(2018{\natexlab{c}})Habernal, Wachsmuth, Gurevych, and
  Stein}]{habernal-etal-2018-name}
Ivan Habernal, Henning Wachsmuth, Iryna Gurevych, and Benno Stein.
  2018{\natexlab{c}}.
\newblock \href {https://doi.org/10.18653/v1/N18-1036} {Before name-calling:
  Dynamics and triggers of ad hominem fallacies in web argumentation}.
\newblock In \emph{Proceedings of the 2018 Conference of the North {A}merican
  Chapter of the Association for Computational Linguistics: Human Language
  Technologies, Volume 1 (Long Papers)}, pages 386--396, New Orleans,
  Louisiana. Association for Computational Linguistics.

\bibitem[{Haddadan et~al.(2019)Haddadan, Cabrio, and
  Villata}]{haddadan-etal-2019-yes}
Shohreh Haddadan, Elena Cabrio, and Serena Villata. 2019.
\newblock \href {https://doi.org/10.18653/v1/P19-1463} {Yes, we can! {{M}}ining
  arguments in 50 years of {US} presidential campaign debates}.
\newblock In \emph{Proceedings of the 57th Annual Meeting of the Association
  for Computational Linguistics}, pages 4684--4690, Florence, Italy.
  Association for Computational Linguistics.

\bibitem[{Haidt and Joseph(2004)}]{haidt2004intuitive}
Jonathan Haidt and Craig Joseph. 2004.
\newblock Intuitive ethics: How innately prepared intuitions generate
  culturally variable virtues.
\newblock \emph{Daedalus}, 133(4):55--66.

\bibitem[{Hamblin(1970)}]{hamblin:1970}
Charles~L. Hamblin. 1970.
\newblock \emph{Fallacies}.
\newblock Methuen, London, UK.

\bibitem[{Hasan and Ng(2014)}]{hasan-ng-2014-taking}
Kazi~Saidul Hasan and Vincent Ng. 2014.
\newblock \href {https://doi.org/10.3115/v1/D14-1083} {Why are you taking this
  stance? {{I}}dentifying and classifying reasons in ideological debates}.
\newblock In \emph{Proceedings of the 2014 Conference on Empirical Methods in
  Natural Language Processing ({EMNLP})}, pages 751--762, Doha, Qatar.
  Association for Computational Linguistics.

\bibitem[{Hawthorne(2002)}]{10.2307/3070991}
John Hawthorne. 2002.
\newblock \href {http://www.jstor.org/stable/3070991} {Deeply contingent a
  priori knowledge}.
\newblock \emph{Philosophy and Phenomenological Research}, 65(2):247--269.

\bibitem[{Hewett et~al.(2019)Hewett, Prakash~Rane, Harlacher, and
  Stede}]{hewett-etal-2019-utility}
Freya Hewett, Roshan Prakash~Rane, Nina Harlacher, and Manfred Stede. 2019.
\newblock \href {https://doi.org/10.18653/v1/W19-4512} {The utility of
  discourse parsing features for predicting argumentation structure}.
\newblock In \emph{Proceedings of the 6th Workshop on Argument Mining}, pages
  98--103, Florence, Italy. Association for Computational Linguistics.

\bibitem[{Hidey and McKeown(2019)}]{hidey-mckeown-2019-fixed}
Christopher Hidey and Kathy McKeown. 2019.
\newblock \href {https://doi.org/10.18653/v1/N19-1174} {Fixed that for you:
  Generating contrastive claims with semantic edits}.
\newblock In \emph{Proceedings of the 2019 Conference of the North {A}merican
  Chapter of the Association for Computational Linguistics: Human Language
  Technologies, Volume 1 (Long and Short Papers)}, pages 1756--1767,
  Minneapolis, Minnesota. Association for Computational Linguistics.

\bibitem[{Hou and Jochim(2017)}]{hou-jochim-2017-argument}
Yufang Hou and Charles Jochim. 2017.
\newblock \href {https://doi.org/10.18653/v1/W17-5107} {Argument relation
  classification using a joint inference model}.
\newblock In \emph{Proceedings of the 4th Workshop on Argument Mining}, pages
  60--66, Copenhagen, Denmark. Association for Computational Linguistics.

\bibitem[{Houlsby et~al.(2019)Houlsby, Giurgiu, Jastrzebski, Morrone,
  De~Laroussilhe, Gesmundo, Attariyan, and Gelly}]{houlsby2019parameter}
Neil Houlsby, Andrei Giurgiu, Stanislaw Jastrzebski, Bruna Morrone, Quentin
  De~Laroussilhe, Andrea Gesmundo, Mona Attariyan, and Sylvain Gelly. 2019.
\newblock Parameter-efficient transfer learning for nlp.
\newblock In \emph{International Conference on Machine Learning}, pages
  2790--2799. PMLR.

\bibitem[{Hu and Liu(2004)}]{hu2004mining}
Minqing Hu and Bing Liu. 2004.
\newblock Mining opinion features in customer reviews.
\newblock \emph{AAAI}, 4(4):755--760.

\bibitem[{Hua et~al.(2019{\natexlab{a}})Hua, Hu, and
  Wang}]{hua-etal-2019-argument-generation}
Xinyu Hua, Zhe Hu, and Lu~Wang. 2019{\natexlab{a}}.
\newblock \href {https://doi.org/10.18653/v1/P19-1255} {Argument generation
  with retrieval, planning, and realization}.
\newblock In \emph{Proceedings of the 57th Annual Meeting of the Association
  for Computational Linguistics}, pages 2661--2672, Florence, Italy.
  Association for Computational Linguistics.

\bibitem[{Hua et~al.(2019{\natexlab{b}})Hua, Nikolov, Badugu, and
  Wang}]{hua-etal-2019-argument}
Xinyu Hua, Mitko Nikolov, Nikhil Badugu, and Lu~Wang. 2019{\natexlab{b}}.
\newblock \href {https://doi.org/10.18653/v1/N19-1219} {Argument mining for
  understanding peer reviews}.
\newblock In \emph{Proceedings of the 2019 Conference of the North {A}merican
  Chapter of the Association for Computational Linguistics: Human Language
  Technologies, Volume 1 (Long and Short Papers)}, pages 2131--2137,
  Minneapolis, Minnesota. Association for Computational Linguistics.

\bibitem[{Hua and Wang(2018)}]{hua-wang-2018-neural}
Xinyu Hua and Lu~Wang. 2018.
\newblock \href {https://doi.org/10.18653/v1/P18-1021} {Neural argument
  generation augmented with externally retrieved evidence}.
\newblock In \emph{Proceedings of the 56th Annual Meeting of the Association
  for Computational Linguistics (Volume 1: Long Papers)}, pages 219--230,
  Melbourne, Australia. Association for Computational Linguistics.

\bibitem[{Hua and Wang(2019)}]{hua-wang-2019-sentence}
Xinyu Hua and Lu~Wang. 2019.
\newblock \href {https://doi.org/10.18653/v1/D19-1055} {Sentence-level content
  planning and style specification for neural text generation}.
\newblock In \emph{Proceedings of the 2019 Conference on Empirical Methods in
  Natural Language Processing and the 9th International Joint Conference on
  Natural Language Processing (EMNLP-IJCNLP)}, pages 591--602, Hong Kong,
  China. Association for Computational Linguistics.

\bibitem[{Huber et~al.(2019)Huber, Toussaint, Roze, Dargnat, and
  Braud}]{huber-etal-2019-aligning}
Laurine Huber, Yannick Toussaint, Charlotte Roze, Mathilde Dargnat, and
  Chlo{\'e} Braud. 2019.
\newblock \href {https://doi.org/10.18653/v1/W19-4504} {Aligning discourse and
  argumentation structures using subtrees and redescription mining}.
\newblock In \emph{Proceedings of the 6th Workshop on Argument Mining}, pages
  35--40, Florence, Italy. Association for Computational Linguistics.

\bibitem[{Ji et~al.(2018)Ji, Wei, Hu, Liu, Zhang, and
  Huang}]{ji-etal-2018-incorporating}
Lu~Ji, Zhongyu Wei, Xiangkun Hu, Yang Liu, Qi~Zhang, and Xuanjing Huang. 2018.
\newblock \href {https://www.aclweb.org/anthology/C18-1314} {Incorporating
  argument-level interactions for persuasion comments evaluation using
  co-attention model}.
\newblock In \emph{Proceedings of the 27th International Conference on
  Computational Linguistics}, pages 3703--3714, Santa Fe, New Mexico, USA.
  Association for Computational Linguistics.

\bibitem[{Ji et~al.(2021)Ji, Pan, Cambria, Marttinen, and
  Philip}]{ji2021survey}
Shaoxiong Ji, Shirui Pan, Erik Cambria, Pekka Marttinen, and S~Yu Philip. 2021.
\newblock A survey on knowledge graphs: Representation, acquisition, and
  applications.
\newblock \emph{IEEE Transactions on Neural Networks and Learning Systems}.

\bibitem[{Jo et~al.(2019)Jo, Visser, Reed, and Hovy}]{jo-etal-2019-cascade}
Yohan Jo, Jacky Visser, Chris Reed, and Eduard Hovy. 2019.
\newblock \href {https://doi.org/10.18653/v1/W19-4502} {A cascade model for
  proposition extraction in argumentation}.
\newblock In \emph{Proceedings of the 6th Workshop on Argument Mining}, pages
  11--24, Florence, Italy. Association for Computational Linguistics.

\bibitem[{Kobbe et~al.(2020{\natexlab{a}})Kobbe, Hulpu{\textcommabelow{s}}, and
  Stuckenschmidt}]{kobbe-etal-2020-unsupervised}
Jonathan Kobbe, Ioana Hulpu{\textcommabelow{s}}, and Heiner Stuckenschmidt.
  2020{\natexlab{a}}.
\newblock \href {https://doi.org/10.18653/v1/2020.emnlp-main.4} {Unsupervised
  stance detection for arguments from consequences}.
\newblock In \emph{Proceedings of the 2020 Conference on Empirical Methods in
  Natural Language Processing (EMNLP)}, pages 50--60, Online. Association for
  Computational Linguistics.

\bibitem[{Kobbe et~al.(2020{\natexlab{b}})Kobbe, Rehbein,
  Hulpu{\textcommabelow{s}}, and Stuckenschmidt}]{kobbe-etal-2020-exploring}
Jonathan Kobbe, Ines Rehbein, Ioana Hulpu{\textcommabelow{s}}, and Heiner
  Stuckenschmidt. 2020{\natexlab{b}}.
\newblock \href {https://www.aclweb.org/anthology/2020.argmining-1.4}
  {Exploring morality in argumentation}.
\newblock In \emph{Proceedings of the 7th Workshop on Argument Mining}, pages
  30--40, Online. Association for Computational Linguistics.

\bibitem[{Kotonya and Toni(2019)}]{kotonya-toni-2019-gradual}
Neema Kotonya and Francesca Toni. 2019.
\newblock \href {https://doi.org/10.18653/v1/W19-4518} {Gradual argumentation
  evaluation for stance aggregation in automated fake news detection}.
\newblock In \emph{Proceedings of the 6th Workshop on Argument Mining}, pages
  156--166, Florence, Italy. Association for Computational Linguistics.

\bibitem[{Kouylekov and Negri(2010)}]{kouylekov2010open}
Milen Kouylekov and Matteo Negri. 2010.
\newblock An open-source package for recognizing textual entailment.
\newblock In \emph{Proceedings of the ACL 2010 System Demonstrations}, pages
  42--47.

\bibitem[{Landis and Koch(1977)}]{landis1977measurement}
J~Richard Landis and Gary~G Koch. 1977.
\newblock The measurement of observer agreement for categorical data.
\newblock \emph{biometrics}, pages 159--174.

\bibitem[{Lauscher et~al.(2018)Lauscher, Glava{\v{s}}, Ponzetto, and
  Eckert}]{lauscher-etal-2018-investigating}
Anne Lauscher, Goran Glava{\v{s}}, Simone~Paolo Ponzetto, and Kai Eckert. 2018.
\newblock \href {https://doi.org/10.18653/v1/D18-1370} {Investigating the role
  of argumentation in the rhetorical analysis of scientific publications with
  neural multi-task learning models}.
\newblock In \emph{Proceedings of the 2018 Conference on Empirical Methods in
  Natural Language Processing}, pages 3326--3338, Brussels, Belgium.
  Association for Computational Linguistics.

\bibitem[{Lauscher et~al.(2021)Lauscher, Ko, Kuhl, Johnson, Jurgens, Cohan, and
  Lo}]{lauscher2021multicite}
Anne Lauscher, Brandon Ko, Bailey Kuhl, Sophie Johnson, David Jurgens, Arman
  Cohan, and Kyle Lo. 2021.
\newblock Multicite: Modeling realistic citations requires moving beyond the
  single-sentence single-label setting.
\newblock \emph{arXiv preprint arXiv:2107.00414}.

\bibitem[{Lauscher et~al.(2020{\natexlab{a}})Lauscher, Majewska, Ribeiro,
  Gurevych, Rozanov, and Glava{\v{s}}}]{lauscher-etal-2020-common}
Anne Lauscher, Olga Majewska, Leonardo F.~R. Ribeiro, Iryna Gurevych, Nikolai
  Rozanov, and Goran Glava{\v{s}}. 2020{\natexlab{a}}.
\newblock \href {https://doi.org/10.18653/v1/2020.deelio-1.5} {Common sense or
  world knowledge? {{I}}nvestigating adapter-based knowledge injection into
  pretrained transformers}.
\newblock In \emph{Proceedings of Deep Learning Inside Out (DeeLIO): The First
  Workshop on Knowledge Extraction and Integration for Deep Learning
  Architectures}, pages 43--49, Online. Association for Computational
  Linguistics.

\bibitem[{Lauscher et~al.(2020{\natexlab{b}})Lauscher, Ng, Napoles, and
  Tetreault}]{lauscher-etal-2020-rhetoric}
Anne Lauscher, Lily Ng, Courtney Napoles, and Joel Tetreault.
  2020{\natexlab{b}}.
\newblock \href {https://doi.org/10.18653/v1/2020.coling-main.402} {Rhetoric,
  logic, and dialectic: Advancing theory-based argument quality assessment in
  natural language processing}.
\newblock In \emph{Proceedings of the 28th International Conference on
  Computational Linguistics}, pages 4563--4574, Barcelona, Spain (Online).
  International Committee on Computational Linguistics.

\bibitem[{Lawrence and Reed(2015)}]{lawrence-reed-2015-combining}
John Lawrence and Chris Reed. 2015.
\newblock \href {https://doi.org/10.3115/v1/W15-0516} {Combining argument
  mining techniques}.
\newblock In \emph{Proceedings of the 2nd Workshop on Argumentation Mining},
  pages 127--136, Denver, CO. Association for Computational Linguistics.

\bibitem[{Lawrence and Reed(2017{\natexlab{a}})}]{lawrence-reed-2017-mining}
John Lawrence and Chris Reed. 2017{\natexlab{a}}.
\newblock \href {https://doi.org/10.18653/v1/W17-5105} {Mining argumentative
  structure from natural language text using automatically generated
  premise-conclusion topic models}.
\newblock In \emph{Proceedings of the 4th Workshop on Argument Mining}, pages
  39--48, Copenhagen, Denmark. Association for Computational Linguistics.

\bibitem[{Lawrence and Reed(2017{\natexlab{b}})}]{lawrence-reed-2017-using}
John Lawrence and Chris Reed. 2017{\natexlab{b}}.
\newblock \href {https://doi.org/10.18653/v1/W17-5114} {Using complex
  argumentative interactions to reconstruct the argumentative structure of
  large-scale debates}.
\newblock In \emph{Proceedings of the 4th Workshop on Argument Mining}, pages
  108--117, Copenhagen, Denmark. Association for Computational Linguistics.

\bibitem[{Lawrence and Reed(2020)}]{lawrence2020argument}
John Lawrence and Chris Reed. 2020.
\newblock Argument mining: A survey.
\newblock \emph{Computational Linguistics}, 45(4):765--818.

\bibitem[{Le et~al.(2018)Le, Nguyen, and Nguyen}]{le-etal-2018-dave}
Dieu~Thu Le, Cam-Tu Nguyen, and Kim~Anh Nguyen. 2018.
\newblock \href {https://doi.org/10.18653/v1/W18-5215} {Dave the debater: a
  retrieval-based and generative argumentative dialogue agent}.
\newblock In \emph{Proceedings of the 5th Workshop on Argument Mining}, pages
  121--130, Brussels, Belgium. Association for Computational Linguistics.

\bibitem[{Levy et~al.(2017)Levy, Gretz, Sznajder, Hummel, Aharonov, and
  Slonim}]{levy-etal-2017-unsupervised}
Ran Levy, Shai Gretz, Benjamin Sznajder, Shay Hummel, Ranit Aharonov, and Noam
  Slonim. 2017.
\newblock \href {https://doi.org/10.18653/v1/W17-5110} {Unsupervised
  corpus{--}wide claim detection}.
\newblock In \emph{Proceedings of the 4th Workshop on Argument Mining}, pages
  79--84, Copenhagen, Denmark. Association for Computational Linguistics.

\bibitem[{Li et~al.(2020)Li, Durmus, and Cardie}]{li-etal-2020-exploring}
Jialu Li, Esin Durmus, and Claire Cardie. 2020.
\newblock \href {https://doi.org/10.18653/v1/2020.emnlp-main.716} {Exploring
  the role of argument structure in online debate persuasion}.
\newblock In \emph{Proceedings of the 2020 Conference on Empirical Methods in
  Natural Language Processing (EMNLP)}, pages 8905--8912, Online. Association
  for Computational Linguistics.

\bibitem[{Liebeck et~al.(2016)Liebeck, Esau, and
  Conrad}]{liebeck-etal-2016-airport}
Matthias Liebeck, Katharina Esau, and Stefan Conrad. 2016.
\newblock \href {https://doi.org/10.18653/v1/W16-2817} {What to do with an
  airport? mining arguments in the {G}erman online participation project
  tempelhofer feld}.
\newblock In \emph{Proceedings of the Third Workshop on Argument Mining
  ({A}rg{M}ining2016)}, pages 144--153, Berlin, Germany. Association for
  Computational Linguistics.

\bibitem[{Liebeck et~al.(2018)Liebeck, Funke, and
  Conrad}]{liebeck-etal-2018-hhu}
Matthias Liebeck, Andreas Funke, and Stefan Conrad. 2018.
\newblock \href {https://doi.org/10.18653/v1/S18-1188} {{HHU} at
  {S}em{E}val-2018 task 12: Analyzing an ensemble-based deep learning approach
  for the argument mining task of choosing the correct warrant}.
\newblock In \emph{Proceedings of The 12th International Workshop on Semantic
  Evaluation}, pages 1114--1119, New Orleans, Louisiana. Association for
  Computational Linguistics.

\bibitem[{Liga(2019)}]{liga-2019-argumentative}
Davide Liga. 2019.
\newblock \href {https://doi.org/10.18653/v1/W19-4511} {Argumentative evidences
  classification and argument scheme detection using tree kernels}.
\newblock In \emph{Proceedings of the 6th Workshop on Argument Mining}, pages
  92--97, Florence, Italy. Association for Computational Linguistics.

\bibitem[{Lin et~al.(2021)Lin, Lee, Qiao, and Ren}]{lin2021common}
Bill~Yuchen Lin, Seyeon Lee, Xiaoyang Qiao, and Xiang Ren. 2021.
\newblock Common sense beyond english: Evaluating and improving multilingual
  language models for commonsense reasoning.
\newblock In \emph{Proceedings of the 59th Annual Meeting of the Association
  for Computational Linguistics and the 11th International Joint Conference on
  Natural Language Processing (Volume 1: Long Papers)}, pages 1274--1287.

\bibitem[{Lin et~al.(2019)Lin, Huang, Huang, and Chen}]{lin-etal-2019-lexicon}
Jian-Fu Lin, Kuo~Yu Huang, Hen-Hsen Huang, and Hsin-Hsi Chen. 2019.
\newblock \href {https://doi.org/10.18653/v1/W19-4508} {Lexicon guided
  attentive neural network model for argument mining}.
\newblock In \emph{Proceedings of the 6th Workshop on Argument Mining}, pages
  67--73, Florence, Italy. Association for Computational Linguistics.

\bibitem[{Lippi and Torroni(2015)}]{lippi2015argument}
Marco Lippi and Paolo Torroni. 2015.
\newblock Argument mining: A machine learning perspective.
\newblock In \emph{International Workshop on Theory and Applications of Formal
  Argumentation}, pages 163--176. Springer.

\bibitem[{Liu(2012)}]{liu2012sentiment}
Bing Liu. 2012.
\newblock Sentiment analysis and opinion mining.
\newblock \emph{Synthesis lectures on human language technologies},
  5(1):1--167.

\bibitem[{Liu et~al.(2008)Liu, Huang, An, and Yu}]{4781139}
Yang Liu, Xiangji Huang, Aijun An, and Xiaohui Yu. 2008.
\newblock \href {https://doi.org/10.1109/ICDM.2008.94} {Modeling and predicting
  the helpfulness of online reviews}.
\newblock In \emph{2008 Eighth IEEE International Conference on Data Mining},
  pages 443--452.

\bibitem[{Lloyd(2007)}]{lloyd:2007}
Keith Lloyd. 2007.
\newblock \href {http://www.jstor.org/stable/20176805} {Rethinking rhetoric
  from an indian perspective: {I}mplications in the ``nyaya sutra''}.
\newblock \emph{Rhetoric Review}, 26(4):365--384.

\bibitem[{Lugini and Litman(2018)}]{lugini-litman-2018-argument}
Luca Lugini and Diane Litman. 2018.
\newblock \href {https://doi.org/10.18653/v1/W18-5208} {Argument component
  classification for classroom discussions}.
\newblock In \emph{Proceedings of the 5th Workshop on Argument Mining}, pages
  57--67, Brussels, Belgium. Association for Computational Linguistics.

\bibitem[{Lukin et~al.(2017)Lukin, Anand, Walker, and
  Whittaker}]{lukin-etal-2017-argument}
Stephanie Lukin, Pranav Anand, Marilyn Walker, and Steve Whittaker. 2017.
\newblock \href {https://www.aclweb.org/anthology/E17-1070} {Argument strength
  is in the eye of the beholder: Audience effects in persuasion}.
\newblock In \emph{Proceedings of the 15th Conference of the {E}uropean Chapter
  of the Association for Computational Linguistics: Volume 1, Long Papers},
  pages 742--753, Valencia, Spain. Association for Computational Linguistics.

\bibitem[{McBurney and Parsons(2021)}]{mcburney:2021}
Peter McBurney and Simon Parsons. 2021.
\newblock Argument schemes and dialogue protocols: {D}oug
  walton{\textquoteright}s legacy in artificial intelligence.
\newblock \emph{IfCoLoG Journal of Logics and their Applications},
  8(1):263--290.

\bibitem[{Mensonides et~al.(2019)Mensonides, Harispe, Montmain, and
  Thireau}]{mensonides-etal-2019-automatic}
Jean-Christophe Mensonides, S{\'e}bastien Harispe, Jacky Montmain, and
  V{\'e}ronique Thireau. 2019.
\newblock \href {https://www.aclweb.org/anthology/W19-7404} {Automatic
  detection and classification of argument components using multi-task deep
  neural network}.
\newblock In \emph{Proceedings of the 3rd International Conference on Natural
  Language and Speech Processing}, pages 25--33, Trento, Italy. Association for
  Computational Linguistics.

\bibitem[{Mikolov et~al.(2013)Mikolov, Sutskever, Chen, Corrado, and
  Dean}]{mikolov2013distributed}
Tomas Mikolov, Ilya Sutskever, Kai Chen, Greg~S Corrado, and Jeff Dean. 2013.
\newblock Distributed representations of words and phrases and their
  compositionality.
\newblock In \emph{Advances in neural information processing systems}, pages
  3111--3119.

\bibitem[{Mo et~al.(2020)Mo, Yunpeng, Li, and Huan}]{mo2020deep}
Wang Mo, Cui Yunpeng, Chen Li, and Li~Huan. 2020.
\newblock A deep learning-based method of argumentative zoning for research
  articles.
\newblock \emph{Data Analysis and Knowledge Discovery}, 4(6):60--68.

\bibitem[{Moens(2018)}]{moens2018argumentation}
Marie-Francine Moens. 2018.
\newblock \href {https://doi.org/10.3233/aac-170025} {Argumentation mining:
  {How} can a machine acquire common sense and world knowledge?}
\newblock \emph{Argument \& Computation}, 9(1):1--14.

\bibitem[{Morio and Fujita(2018)}]{morio-fujita-2018-end}
Gaku Morio and Katsuhide Fujita. 2018.
\newblock \href {https://doi.org/10.18653/v1/W18-5202} {End-to-end argument
  mining for discussion threads based on parallel constrained pointer
  architecture}.
\newblock In \emph{Proceedings of the 5th Workshop on Argument Mining}, pages
  11--21, Brussels, Belgium. Association for Computational Linguistics.

\bibitem[{Morio et~al.(2020)Morio, Ozaki, Morishita, Koreeda, and
  Yanai}]{morio-etal-2020-towards}
Gaku Morio, Hiroaki Ozaki, Terufumi Morishita, Yuta Koreeda, and Kohsuke Yanai.
  2020.
\newblock \href {https://doi.org/10.18653/v1/2020.acl-main.298} {Towards better
  non-tree argument mining: Proposition-level biaffine parsing with
  task-specific parameterization}.
\newblock In \emph{Proceedings of the 58th Annual Meeting of the Association
  for Computational Linguistics}, pages 3259--3266, Online. Association for
  Computational Linguistics.

\bibitem[{Niculae et~al.(2017)Niculae, Park, and
  Cardie}]{niculae-etal-2017-argument}
Vlad Niculae, Joonsuk Park, and Claire Cardie. 2017.
\newblock \href {https://doi.org/10.18653/v1/P17-1091} {Argument mining with
  structured {SVM}s and {RNN}s}.
\newblock In \emph{Proceedings of the 55th Annual Meeting of the Association
  for Computational Linguistics (Volume 1: Long Papers)}, pages 985--995,
  Vancouver, Canada. Association for Computational Linguistics.

\bibitem[{Niven and Kao(2019)}]{niven-kao-2019-probing}
Timothy Niven and Hung-Yu Kao. 2019.
\newblock \href {https://doi.org/10.18653/v1/P19-1459} {Probing neural network
  comprehension of natural language arguments}.
\newblock In \emph{Proceedings of the 57th Annual Meeting of the Association
  for Computational Linguistics}, pages 4658--4664, Florence, Italy.
  Association for Computational Linguistics.

\bibitem[{Ong et~al.(2014)Ong, Litman, and
  Brusilovsky}]{ong-etal-2014-ontology}
Nathan Ong, Diane Litman, and Alexandra Brusilovsky. 2014.
\newblock \href {https://doi.org/10.3115/v1/W14-2104} {Ontology-based argument
  mining and automatic essay scoring}.
\newblock In \emph{Proceedings of the First Workshop on Argumentation Mining},
  pages 24--28, Baltimore, Maryland. Association for Computational Linguistics.

\bibitem[{Opitz and Frank(2019)}]{opitz-frank-2019-dissecting}
Juri Opitz and Anette Frank. 2019.
\newblock \href {https://doi.org/10.18653/v1/W19-4503} {Dissecting content and
  context in argumentative relation analysis}.
\newblock In \emph{Proceedings of the 6th Workshop on Argument Mining}, pages
  25--34, Florence, Italy. Association for Computational Linguistics.

\bibitem[{Passon et~al.(2018)Passon, Lippi, Serra, and
  Tasso}]{passon-etal-2018-predicting}
Marco Passon, Marco Lippi, Giuseppe Serra, and Carlo Tasso. 2018.
\newblock \href {https://doi.org/10.18653/v1/W18-5205} {Predicting the
  usefulness of {A}mazon reviews using off-the-shelf argumentation mining}.
\newblock In \emph{Proceedings of the 5th Workshop on Argument Mining}, pages
  35--39, Brussels, Belgium. Association for Computational Linguistics.

\bibitem[{Paul et~al.(2020)Paul, Opitz, Becker, Kobbe, Hirst, and
  Frank}]{paul2020argumentative}
Debjit Paul, Juri Opitz, Maria Becker, Jonathan Kobbe, Graeme Hirst, and Anette
  Frank. 2020.
\newblock Argumentative relation classification with background knowledge.
\newblock In \emph{Computational Models of Argument}, pages 319--330. IOS
  Press.

\bibitem[{Peldszus and Stede(2015)}]{peldszus-stede-2015-joint}
Andreas Peldszus and Manfred Stede. 2015.
\newblock \href {https://doi.org/10.18653/v1/D15-1110} {Joint prediction in
  {MST}-style discourse parsing for argumentation mining}.
\newblock In \emph{Proceedings of the 2015 Conference on Empirical Methods in
  Natural Language Processing}, pages 938--948, Lisbon, Portugal. Association
  for Computational Linguistics.

\bibitem[{Pennington et~al.(2014)Pennington, Socher, and
  Manning}]{pennington-etal-2014-glove}
Jeffrey Pennington, Richard Socher, and Christopher Manning. 2014.
\newblock \href {https://doi.org/10.3115/v1/D14-1162} {{G}lo{V}e: Global
  vectors for word representation}.
\newblock In \emph{Proceedings of the 2014 Conference on Empirical Methods in
  Natural Language Processing ({EMNLP})}, pages 1532--1543, Doha, Qatar.
  Association for Computational Linguistics.

\bibitem[{Persing et~al.(2010)Persing, Davis, and
  Ng}]{persing-etal-2010-modeling}
Isaac Persing, Alan Davis, and Vincent Ng. 2010.
\newblock \href {https://www.aclweb.org/anthology/D10-1023} {Modeling
  organization in student essays}.
\newblock In \emph{Proceedings of the 2010 Conference on Empirical Methods in
  Natural Language Processing}, pages 229--239, Cambridge, MA. Association for
  Computational Linguistics.

\bibitem[{Persing and Ng(2013)}]{persing-ng-2013-clarity}
Isaac Persing and Vincent Ng. 2013.
\newblock \href {https://www.aclweb.org/anthology/P13-1026} {Modeling thesis
  clarity in student essays}.
\newblock In \emph{Proceedings of the 51st Annual Meeting of the Association
  for Computational Linguistics (Volume 1: Long Papers)}, pages 260--269,
  Sofia, Bulgaria. Association for Computational Linguistics.

\bibitem[{Persing and Ng(2014)}]{persing-ng-2014-modeling}
Isaac Persing and Vincent Ng. 2014.
\newblock \href {https://doi.org/10.3115/v1/P14-1144} {Modeling prompt
  adherence in student essays}.
\newblock In \emph{Proceedings of the 52nd Annual Meeting of the Association
  for Computational Linguistics (Volume 1: Long Papers)}, pages 1534--1543,
  Baltimore, Maryland. Association for Computational Linguistics.

\bibitem[{Persing and Ng(2015)}]{persing-ng-2015-modeling}
Isaac Persing and Vincent Ng. 2015.
\newblock \href {https://doi.org/10.3115/v1/P15-1053} {Modeling argument
  strength in student essays}.
\newblock In \emph{Proceedings of the 53rd Annual Meeting of the Association
  for Computational Linguistics and the 7th International Joint Conference on
  Natural Language Processing (Volume 1: Long Papers)}, pages 543--552,
  Beijing, China. Association for Computational Linguistics.

\bibitem[{Persing and Ng(2016{\natexlab{a}})}]{persing-ng-2016-end}
Isaac Persing and Vincent Ng. 2016{\natexlab{a}}.
\newblock \href {https://doi.org/10.18653/v1/N16-1164} {End-to-end
  argumentation mining in student essays}.
\newblock In \emph{Proceedings of the 2016 Conference of the North {A}merican
  Chapter of the Association for Computational Linguistics: Human Language
  Technologies}, pages 1384--1394, San Diego, California. Association for
  Computational Linguistics.

\bibitem[{Persing and Ng(2016{\natexlab{b}})}]{persing-ng-2016-modeling}
Isaac Persing and Vincent Ng. 2016{\natexlab{b}}.
\newblock \href {https://doi.org/10.18653/v1/P16-1205} {Modeling stance in
  student essays}.
\newblock In \emph{Proceedings of the 54th Annual Meeting of the Association
  for Computational Linguistics (Volume 1: Long Papers)}, pages 2174--2184,
  Berlin, Germany. Association for Computational Linguistics.

\bibitem[{Persing and Ng(2017)}]{ijcai2017-570}
Isaac Persing and Vincent Ng. 2017.
\newblock \href {https://doi.org/10.24963/ijcai.2017/570} {Why can't you
  convince me? {{M}}odeling weaknesses in unpersuasive arguments}.
\newblock In \emph{Proceedings of the Twenty-Sixth International Joint
  Conference on Artificial Intelligence, {IJCAI-17}}, pages 4082--4088.

\bibitem[{Persing and Ng(2020)}]{persing-ng-2020-unsupervised}
Isaac Persing and Vincent Ng. 2020.
\newblock \href {https://www.aclweb.org/anthology/2020.lrec-1.839}
  {Unsupervised argumentation mining in student essays}.
\newblock In \emph{Proceedings of the 12th Language Resources and Evaluation
  Conference}, pages 6795--6803, Marseille, France. European Language Resources
  Association.

\bibitem[{Petasis(2019)}]{petasis-2019-segmentation}
Georgios Petasis. 2019.
\newblock \href {https://doi.org/10.18653/v1/W19-4501} {Segmentation of
  argumentative texts with contextualised word representations}.
\newblock In \emph{Proceedings of the 6th Workshop on Argument Mining}, pages
  1--10, Florence, Italy. Association for Computational Linguistics.

\bibitem[{Plato(ca. 400 B.C.E.)}]{plato2019theaetetus}
Plato. ca. 400 B.C.E.
\newblock \emph{Theaetetus}, 2014 edition.
\newblock Oxford University Press.
\newblock Translated by John McDowell.

\bibitem[{Ponti et~al.(2022)Ponti, Sordoni, Bengio, and
  Reddy}]{ponti2022combining}
Edoardo~M Ponti, Alessandro Sordoni, Yoshua Bengio, and Siva Reddy. 2022.
\newblock Combining modular skills in multitask learning.
\newblock \emph{arXiv e-prints}, pages arXiv--2202.

\bibitem[{Ponti et~al.(2020)Ponti, Glava{\v{s}}, Majewska, Liu, Vuli{\'c}, and
  Korhonen}]{ponti2020xcopa}
Edoardo~Maria Ponti, Goran Glava{\v{s}}, Olga Majewska, Qianchu Liu, Ivan
  Vuli{\'c}, and Anna Korhonen. 2020.
\newblock Xcopa: A multilingual dataset for causal commonsense reasoning.
\newblock In \emph{Proceedings of the 2020 Conference on Empirical Methods in
  Natural Language Processing (EMNLP)}, pages 2362--2376.

\bibitem[{Porco and Goldwasser(2020)}]{porco-goldwasser-2020-predicting}
Aldo Porco and Dan Goldwasser. 2020.
\newblock \href {https://doi.org/10.18653/v1/2020.coling-main.35} {Predicting
  stance change using modular architectures}.
\newblock In \emph{Proceedings of the 28th International Conference on
  Computational Linguistics}, pages 396--406, Barcelona, Spain (Online).
  International Committee on Computational Linguistics.

\bibitem[{Potash et~al.(2017{\natexlab{a}})Potash, Bhattacharya, and
  Rumshisky}]{potash-etal-2017-length}
Peter Potash, Robin Bhattacharya, and Anna Rumshisky. 2017{\natexlab{a}}.
\newblock \href {https://www.aclweb.org/anthology/I17-1035} {Length,
  interchangeability, and external knowledge: Observations from predicting
  argument convincingness}.
\newblock In \emph{Proceedings of the Eighth International Joint Conference on
  Natural Language Processing (Volume 1: Long Papers)}, pages 342--351, Taipei,
  Taiwan. Asian Federation of Natural Language Processing.

\bibitem[{Potash et~al.(2019)Potash, Ferguson, and
  Hazen}]{potash-etal-2019-ranking}
Peter Potash, Adam Ferguson, and Timothy~J. Hazen. 2019.
\newblock \href {https://doi.org/10.18653/v1/W19-4517} {Ranking passages for
  argument convincingness}.
\newblock In \emph{Proceedings of the 6th Workshop on Argument Mining}, pages
  146--155, Florence, Italy. Association for Computational Linguistics.

\bibitem[{Potash et~al.(2017{\natexlab{b}})Potash, Romanov, and
  Rumshisky}]{potash-etal-2017-heres}
Peter Potash, Alexey Romanov, and Anna Rumshisky. 2017{\natexlab{b}}.
\newblock \href {https://doi.org/10.18653/v1/D17-1143} {Here{'}s my point:
  Joint pointer architecture for argument mining}.
\newblock In \emph{Proceedings of the 2017 Conference on Empirical Methods in
  Natural Language Processing}, pages 1364--1373, Copenhagen, Denmark.
  Association for Computational Linguistics.

\bibitem[{Potthast et~al.(2019)Potthast, Gienapp, Euchner, Heilenk\"{o}tter,
  Weidmann, Wachsmuth, Stein, and Hagen}]{10.1145/3331184.3331327}
Martin Potthast, Lukas Gienapp, Florian Euchner, Nick Heilenk\"{o}tter, Nico
  Weidmann, Henning Wachsmuth, Benno Stein, and Matthias Hagen. 2019.
\newblock \href {https://doi.org/10.1145/3331184.3331327} {Argument search:
  Assessing argument relevance}.
\newblock In \emph{Proceedings of the 42nd International ACM SIGIR Conference
  on Research and Development in Information Retrieval}, SIGIR'19, page
  1117–1120, New York, NY, USA. Association for Computing Machinery.

\bibitem[{Rajani et~al.(2019)Rajani, McCann, Xiong, and
  Socher}]{rajani2019explain}
Nazneen~Fatema Rajani, Bryan McCann, Caiming Xiong, and Richard Socher. 2019.
\newblock Explain yourself! leveraging language models for commonsense
  reasoning.
\newblock In \emph{Proceedings of the 57th Annual Meeting of the Association
  for Computational Linguistics}, pages 4932--4942.

\bibitem[{Rajendran et~al.(2018{\natexlab{a}})Rajendran, Bollegala, and
  Parsons}]{rajendran-etal-2018-something}
Pavithra Rajendran, Danushka Bollegala, and Simon Parsons. 2018{\natexlab{a}}.
\newblock \href {https://doi.org/10.18653/v1/N18-2005} {Is something better
  than nothing? {{A}}utomatically predicting stance-based arguments using deep
  learning and small labelled dataset}.
\newblock In \emph{Proceedings of the 2018 Conference of the North {A}merican
  Chapter of the Association for Computational Linguistics: Human Language
  Technologies, Volume 2 (Short Papers)}, pages 28--34, New Orleans, Louisiana.
  Association for Computational Linguistics.

\bibitem[{Rajendran et~al.(2018{\natexlab{b}})Rajendran, Bollegala, and
  Parsons}]{rajendran-etal-2018-sentiment}
Pavithra Rajendran, Danushka Bollegala, and Simon Parsons. 2018{\natexlab{b}}.
\newblock \href {https://www.aclweb.org/anthology/L18-1099}
  {Sentiment-stance-specificity ({SSS}) dataset: Identifying support-based
  entailment among opinions.}
\newblock In \emph{Proceedings of the Eleventh International Conference on
  Language Resources and Evaluation ({LREC} 2018)}, Miyazaki, Japan. European
  Language Resources Association (ELRA).

\bibitem[{Ranade et~al.(2013)Ranade, Sangal, and
  Mamidi}]{ranade-etal-2013-stance}
Sarvesh Ranade, Rajeev Sangal, and Radhika Mamidi. 2013.
\newblock \href {https://www.aclweb.org/anthology/W13-4008} {Stance
  classification in online debates by recognizing users{'} intentions}.
\newblock In \emph{Proceedings of the {SIGDIAL} 2013 Conference}, pages 61--69,
  Metz, France. Association for Computational Linguistics.

\bibitem[{Reimers et~al.(2019)Reimers, Schiller, Beck, Daxenberger, Stab, and
  Gurevych}]{reimers-etal-2019-classification}
Nils Reimers, Benjamin Schiller, Tilman Beck, Johannes Daxenberger, Christian
  Stab, and Iryna Gurevych. 2019.
\newblock \href {https://doi.org/10.18653/v1/P19-1054} {Classification and
  clustering of arguments with contextualized word embeddings}.
\newblock In \emph{Proceedings of the 57th Annual Meeting of the Association
  for Computational Linguistics}, pages 567--578, Florence, Italy. Association
  for Computational Linguistics.

\bibitem[{Reisert et~al.(2015)Reisert, Inoue, Okazaki, and
  Inui}]{reisert-etal-2015-computational}
Paul Reisert, Naoya Inoue, Naoaki Okazaki, and Kentaro Inui. 2015.
\newblock \href {https://doi.org/10.3115/v1/W15-0507} {A computational approach
  for generating toulmin model argumentation}.
\newblock In \emph{Proceedings of the 2nd Workshop on Argumentation Mining},
  pages 45--55, Denver, CO. Association for Computational Linguistics.

\bibitem[{Rinott et~al.(2015)Rinott, Dankin, Alzate~Perez, Khapra, Aharoni, and
  Slonim}]{rinott-etal-2015-show}
Ruty Rinott, Lena Dankin, Carlos Alzate~Perez, Mitesh~M. Khapra, Ehud Aharoni,
  and Noam Slonim. 2015.
\newblock \href {https://doi.org/10.18653/v1/D15-1050} {Show me your evidence -
  an automatic method for context dependent evidence detection}.
\newblock In \emph{Proceedings of the 2015 Conference on Empirical Methods in
  Natural Language Processing}, pages 440--450, Lisbon, Portugal. Association
  for Computational Linguistics.

\bibitem[{Saint-Dizier(2017)}]{saint-dizier-2017-using}
Patrick Saint-Dizier. 2017.
\newblock \href {https://doi.org/10.18653/v1/W17-5111} {Using
  question-answering techniques to implement a knowledge-driven argument mining
  approach}.
\newblock In \emph{Proceedings of the 4th Workshop on Argument Mining}, pages
  85--90, Copenhagen, Denmark. Association for Computational Linguistics.

\bibitem[{Sap et~al.(2020)Sap, Shwartz, Bosselut, Choi, and
  Roth}]{sap2020commonsense}
Maarten Sap, Vered Shwartz, Antoine Bosselut, Yejin Choi, and Dan Roth. 2020.
\newblock Commonsense reasoning for natural language processing.
\newblock In \emph{Proceedings of the 58th Annual Meeting of the Association
  for Computational Linguistics: Tutorial Abstracts}, pages 27--33.

\bibitem[{Sato et~al.(2015)Sato, Yanai, Miyoshi, Yanase, Iwayama, Sun, and
  Niwa}]{sato-etal-2015-end}
Misa Sato, Kohsuke Yanai, Toshinori Miyoshi, Toshihiko Yanase, Makoto Iwayama,
  Qinghua Sun, and Yoshiki Niwa. 2015.
\newblock \href {https://doi.org/10.3115/v1/P15-4019} {End-to-end argument
  generation system in debating}.
\newblock In \emph{Proceedings of {ACL}-{IJCNLP} 2015 System Demonstrations},
  pages 109--114, Beijing, China. Association for Computational Linguistics and
  The Asian Federation of Natural Language Processing.

\bibitem[{Schaefer and Stede(2021)}]{schaefer2021argument}
Robin Schaefer and Manfred Stede. 2021.
\newblock Argument mining on twitter: A survey.
\newblock \emph{it-Information Technology}, 63(1):45--58.

\bibitem[{Schiller et~al.(2021)Schiller, Daxenberger, and
  Gurevych}]{schiller-etal-2021-aspect}
Benjamin Schiller, Johannes Daxenberger, and Iryna Gurevych. 2021.
\newblock \href {https://www.aclweb.org/anthology/2021.naacl-main.34}
  {Aspect-controlled neural argument generation}.
\newblock In \emph{Proceedings of the 2021 Conference of the North American
  Chapter of the Association for Computational Linguistics: Human Language
  Technologies}, pages 380--396, Online. Association for Computational
  Linguistics.

\bibitem[{Schulz et~al.(2018)Schulz, Eger, Daxenberger, Kahse, and
  Gurevych}]{schulz-etal-2018-multi}
Claudia Schulz, Steffen Eger, Johannes Daxenberger, Tobias Kahse, and Iryna
  Gurevych. 2018.
\newblock \href {https://doi.org/10.18653/v1/N18-2006} {Multi-task learning for
  argumentation mining in low-resource settings}.
\newblock In \emph{Proceedings of the 2018 Conference of the North {A}merican
  Chapter of the Association for Computational Linguistics: Human Language
  Technologies, Volume 2 (Short Papers)}, pages 35--41, New Orleans, Louisiana.
  Association for Computational Linguistics.

\bibitem[{Scialom et~al.(2020)Scialom, Tekiro{\u{g}}lu, Staiano, and
  Guerini}]{scialom-etal-2020-toward}
Thomas Scialom, Serra~Sinem Tekiro{\u{g}}lu, Jacopo Staiano, and Marco Guerini.
  2020.
\newblock \href {https://doi.org/10.18653/v1/2020.findings-emnlp.238} {Toward
  stance-based personas for opinionated dialogues}.
\newblock In \emph{Findings of the Association for Computational Linguistics:
  EMNLP 2020}, pages 2625--2635, Online. Association for Computational
  Linguistics.

\bibitem[{Shnarch et~al.(2018)Shnarch, Alzate, Dankin, Gleize, Hou, Choshen,
  Aharonov, and Slonim}]{shnarch-etal-2018-will}
Eyal Shnarch, Carlos Alzate, Lena Dankin, Martin Gleize, Yufang Hou, Leshem
  Choshen, Ranit Aharonov, and Noam Slonim. 2018.
\newblock \href {https://doi.org/10.18653/v1/P18-2095} {Will it blend? blending
  weak and strong labeled data in a neural network for argumentation mining}.
\newblock In \emph{Proceedings of the 56th Annual Meeting of the Association
  for Computational Linguistics (Volume 2: Short Papers)}, pages 599--605,
  Melbourne, Australia. Association for Computational Linguistics.

\bibitem[{Shnarch et~al.(2017)Shnarch, Levy, Raykar, and
  Slonim}]{shnarch-etal-2017-grasp}
Eyal Shnarch, Ran Levy, Vikas Raykar, and Noam Slonim. 2017.
\newblock \href {https://doi.org/10.18653/v1/D17-1140} {{GRASP}: Rich patterns
  for argumentation mining}.
\newblock In \emph{Proceedings of the 2017 Conference on Empirical Methods in
  Natural Language Processing}, pages 1345--1350, Copenhagen, Denmark.
  Association for Computational Linguistics.

\bibitem[{Simpson and Gurevych(2018)}]{simpson-gurevych-2018-finding}
Edwin Simpson and Iryna Gurevych. 2018.
\newblock \href {https://doi.org/10.1162/tacl_a_00026} {Finding convincing
  arguments using scalable {B}ayesian preference learning}.
\newblock \emph{Transactions of the Association for Computational Linguistics},
  6:357--371.

\bibitem[{Sirrianni et~al.(2020)Sirrianni, Liu, and
  Adams}]{sirrianni-etal-2020-agreement}
Joseph Sirrianni, Xiaoqing Liu, and Douglas Adams. 2020.
\newblock \href {https://doi.org/10.18653/v1/2020.acl-main.509} {Agreement
  prediction of arguments in cyber argumentation for detecting stance polarity
  and intensity}.
\newblock In \emph{Proceedings of the 58th Annual Meeting of the Association
  for Computational Linguistics}, pages 5746--5758, Online. Association for
  Computational Linguistics.

\bibitem[{Skitalinskaya et~al.(2021)Skitalinskaya, Klaff, and
  Wachsmuth}]{skitalinskaya-etal-2021-learning}
Gabriella Skitalinskaya, Jonas Klaff, and Henning Wachsmuth. 2021.
\newblock \href {https://www.aclweb.org/anthology/2021.eacl-main.147} {Learning
  from revisions: Quality assessment of claims in argumentation at scale}.
\newblock In \emph{Proceedings of the 16th Conference of the European Chapter
  of the Association for Computational Linguistics: Main Volume}, pages
  1718--1729, Online. Association for Computational Linguistics.

\bibitem[{Slonim(2018)}]{slonim2018project}
Noam Slonim. 2018.
\newblock Project debater.
\newblock In \emph{COMMA}, page~4.

\bibitem[{Sobhani et~al.(2015)Sobhani, Inkpen, and
  Matwin}]{sobhani-etal-2015-argumentation}
Parinaz Sobhani, Diana Inkpen, and Stan Matwin. 2015.
\newblock \href {https://doi.org/10.3115/v1/W15-0509} {From argumentation
  mining to stance classification}.
\newblock In \emph{Proceedings of the 2nd Workshop on Argumentation Mining},
  pages 67--77, Denver, CO. Association for Computational Linguistics.

\bibitem[{Sobhani et~al.(2017)Sobhani, Inkpen, and
  Zhu}]{sobhani-etal-2017-dataset}
Parinaz Sobhani, Diana Inkpen, and Xiaodan Zhu. 2017.
\newblock \href {https://www.aclweb.org/anthology/E17-2088} {A dataset for
  multi-target stance detection}.
\newblock In \emph{Proceedings of the 15th Conference of the {E}uropean Chapter
  of the Association for Computational Linguistics: Volume 2, Short Papers},
  pages 551--557, Valencia, Spain. Association for Computational Linguistics.

\bibitem[{Socher et~al.(2013)Socher, Perelygin, Wu, Chuang, Manning, Ng, and
  Potts}]{socher-etal-2013-recursive}
Richard Socher, Alex Perelygin, Jean Wu, Jason Chuang, Christopher~D. Manning,
  Andrew Ng, and Christopher Potts. 2013.
\newblock \href {https://www.aclweb.org/anthology/D13-1170} {Recursive deep
  models for semantic compositionality over a sentiment treebank}.
\newblock In \emph{Proceedings of the 2013 Conference on Empirical Methods in
  Natural Language Processing}, pages 1631--1642, Seattle, Washington, USA.
  Association for Computational Linguistics.

\bibitem[{Somasundaran and Wiebe(2010)}]{somasundaran-wiebe-2010-recognizing}
Swapna Somasundaran and Janyce Wiebe. 2010.
\newblock \href {https://www.aclweb.org/anthology/W10-0214} {Recognizing
  stances in ideological on-line debates}.
\newblock In \emph{Proceedings of the {NAACL} {HLT} 2010 Workshop on
  Computational Approaches to Analysis and Generation of Emotion in Text},
  pages 116--124, Los Angeles, CA. Association for Computational Linguistics.

\bibitem[{Song et~al.(2014)Song, Heilman, Beigman~Klebanov, and
  Deane}]{song-etal-2014-applying}
Yi~Song, Michael Heilman, Beata Beigman~Klebanov, and Paul Deane. 2014.
\newblock \href {https://doi.org/10.3115/v1/W14-2110} {Applying argumentation
  schemes for essay scoring}.
\newblock In \emph{Proceedings of the First Workshop on Argumentation Mining},
  pages 69--78, Baltimore, Maryland. Association for Computational Linguistics.

\bibitem[{Splieth{\"o}ver et~al.(2019)Splieth{\"o}ver, Klaff, and
  Heuer}]{spliethover-etal-2019-worth}
Maximilian Splieth{\"o}ver, Jonas Klaff, and Hendrik Heuer. 2019.
\newblock \href {https://doi.org/10.18653/v1/W19-4509} {Is it worth the
  attention? {{A}} comparative evaluation of attention layers for argument unit
  segmentation}.
\newblock In \emph{Proceedings of the 6th Workshop on Argument Mining}, pages
  74--82, Florence, Italy. Association for Computational Linguistics.

\bibitem[{Stab et~al.(2018{\natexlab{a}})Stab, Daxenberger, Stahlhut, Miller,
  Schiller, Tauchmann, Eger, and Gurevych}]{stab2018argumentext}
Christian Stab, Johannes Daxenberger, Chris Stahlhut, Tristan Miller, Benjamin
  Schiller, Christopher Tauchmann, Steffen Eger, and Iryna Gurevych.
  2018{\natexlab{a}}.
\newblock Argumentext: Searching for arguments in heterogeneous sources.
\newblock In \emph{Proceedings of the 2018 conference of the North American
  chapter of the association for computational linguistics: demonstrations},
  pages 21--25.

\bibitem[{Stab and Gurevych(2014)}]{stab-gurevych-2014-identifying}
Christian Stab and Iryna Gurevych. 2014.
\newblock \href {https://doi.org/10.3115/v1/D14-1006} {Identifying
  argumentative discourse structures in persuasive essays}.
\newblock In \emph{Proceedings of the 2014 Conference on Empirical Methods in
  Natural Language Processing ({EMNLP})}, pages 46--56, Doha, Qatar.
  Association for Computational Linguistics.

\bibitem[{Stab and Gurevych(2016)}]{stab-gurevych-2016-recognizing}
Christian Stab and Iryna Gurevych. 2016.
\newblock \href {https://doi.org/10.18653/v1/W16-2813} {Recognizing the absence
  of opposing arguments in persuasive essays}.
\newblock In \emph{Proceedings of the Third Workshop on Argument Mining
  ({A}rg{M}ining2016)}, pages 113--118, Berlin, Germany. Association for
  Computational Linguistics.

\bibitem[{Stab and Gurevych(2017{\natexlab{a}})}]{stab2017parsing}
Christian Stab and Iryna Gurevych. 2017{\natexlab{a}}.
\newblock Parsing argumentation structures in persuasive essays.
\newblock \emph{Computational Linguistics}, 43(3):619--659.

\bibitem[{Stab and
  Gurevych(2017{\natexlab{b}})}]{stab-gurevych-2017-recognizing}
Christian Stab and Iryna Gurevych. 2017{\natexlab{b}}.
\newblock \href {https://www.aclweb.org/anthology/E17-1092} {Recognizing
  insufficiently supported arguments in argumentative essays}.
\newblock In \emph{Proceedings of the 15th Conference of the {E}uropean Chapter
  of the Association for Computational Linguistics: Volume 1, Long Papers},
  pages 980--990, Valencia, Spain. Association for Computational Linguistics.

\bibitem[{Stab et~al.(2018{\natexlab{b}})Stab, Miller, Schiller, Rai, and
  Gurevych}]{stab-etal-2018-cross}
Christian Stab, Tristan Miller, Benjamin Schiller, Pranav Rai, and Iryna
  Gurevych. 2018{\natexlab{b}}.
\newblock \href {https://doi.org/10.18653/v1/D18-1402} {Cross-topic argument
  mining from heterogeneous sources}.
\newblock In \emph{Proceedings of the 2018 Conference on Empirical Methods in
  Natural Language Processing}, pages 3664--3674, Brussels, Belgium.
  Association for Computational Linguistics.

\bibitem[{Stede and Schneider(2018)}]{stede2018argumentation}
Manfred Stede and Jodi Schneider. 2018.
\newblock Argumentation mining.
\newblock \emph{Synthesis Lectures on Human Language Technologies},
  11(2):1--191.

\bibitem[{Sui et~al.(2018)Sui, Chao, and Luo}]{sui-etal-2018-joker}
Guobin Sui, Wenhan Chao, and Zhunchen Luo. 2018.
\newblock \href {https://doi.org/10.18653/v1/S18-1191} {Joker at
  {S}em{E}val-2018 task 12: The argument reasoning comprehension with neural
  attention}.
\newblock In \emph{Proceedings of The 12th International Workshop on Semantic
  Evaluation}, pages 1129--1132, New Orleans, Louisiana. Association for
  Computational Linguistics.

\bibitem[{Sun et~al.(2018)Sun, Wang, Zhu, and Zhou}]{sun-etal-2018-stance}
Qingying Sun, Zhongqing Wang, Qiaoming Zhu, and Guodong Zhou. 2018.
\newblock \href {https://www.aclweb.org/anthology/C18-1203} {Stance detection
  with hierarchical attention network}.
\newblock In \emph{Proceedings of the 27th International Conference on
  Computational Linguistics}, pages 2399--2409, Santa Fe, New Mexico, USA.
  Association for Computational Linguistics.

\bibitem[{Syed et~al.(2020)Syed, El~Baff, Kiesel, Al~Khatib, Stein, and
  Potthast}]{syed-etal-2020-news}
Shahbaz Syed, Roxanne El~Baff, Johannes Kiesel, Khalid Al~Khatib, Benno Stein,
  and Martin Potthast. 2020.
\newblock \href {https://doi.org/10.18653/v1/2020.coling-main.470} {News
  editorials: Towards summarizing long argumentative texts}.
\newblock In \emph{Proceedings of the 28th International Conference on
  Computational Linguistics}, pages 5384--5396, Barcelona, Spain (Online).
  International Committee on Computational Linguistics.

\bibitem[{Tan et~al.(2016)Tan, Niculae, Danescu-Niculescu-Mizil, and
  Lee}]{10.1145/2872427.2883081}
Chenhao Tan, Vlad Niculae, Cristian Danescu-Niculescu-Mizil, and Lillian Lee.
  2016.
\newblock \href {https://doi.org/10.1145/2872427.2883081} {Winning arguments:
  Interaction dynamics and persuasion strategies in good-faith online
  discussions}.
\newblock In \emph{Proceedings of the 25th International Conference on World
  Wide Web}, WWW '16, page 613–624, Republic and Canton of Geneva, CHE.
  International World Wide Web Conferences Steering Committee.

\bibitem[{Tausczik and Pennebaker(2010)}]{tausczik2010psychological}
Yla~R Tausczik and James~W Pennebaker. 2010.
\newblock The psychological meaning of words: Liwc and computerized text
  analysis methods.
\newblock \emph{Journal of language and social psychology}, 29(1):24--54.

\bibitem[{Teufel et~al.(1999)Teufel, Carletta, and
  Moens}]{teufel1999argumentative}
Simone Teufel, Jean Carletta, and Marc Moens. 1999.
\newblock \href {https://www.aclweb.org/anthology/E99-1015} {An annotation
  scheme for discourse-level argumentation in research articles}.
\newblock In \emph{Ninth Conference of the {E}uropean Chapter of the
  Association for Computational Linguistics}, pages 110--117, Bergen, Norway.
  Association for Computational Linguistics.

\bibitem[{Teufel et~al.(2009)Teufel, Siddharthan, and
  Batchelor}]{teufel2009towards}
Simone Teufel, Advaith Siddharthan, and Colin Batchelor. 2009.
\newblock Towards domain-independent argumentative zoning: Evidence from
  chemistry and computational linguistics.
\newblock In \emph{Proceedings of the 2009 conference on empirical methods in
  natural language processing}, pages 1493--1502.

\bibitem[{Tian et~al.(2018)Tian, Lan, and Wu}]{tian-etal-2018-ecnu}
Junfeng Tian, Man Lan, and Yuanbin Wu. 2018.
\newblock \href {https://doi.org/10.18653/v1/S18-1184} {{ECNU} at
  {S}em{E}val-2018 task 12: An end-to-end attention-based neural network for
  the argument reasoning comprehension task}.
\newblock In \emph{Proceedings of The 12th International Workshop on Semantic
  Evaluation}, pages 1094--1098, New Orleans, Louisiana. Association for
  Computational Linguistics.

\bibitem[{Toledo et~al.(2019)Toledo, Gretz, Cohen-Karlik, Friedman, Venezian,
  Lahav, Jacovi, Aharonov, and Slonim}]{toledo-etal-2019-automatic}
Assaf Toledo, Shai Gretz, Edo Cohen-Karlik, Roni Friedman, Elad Venezian, Dan
  Lahav, Michal Jacovi, Ranit Aharonov, and Noam Slonim. 2019.
\newblock \href {https://doi.org/10.18653/v1/D19-1564} {Automatic argument
  quality assessment - new datasets and methods}.
\newblock In \emph{Proceedings of the 2019 Conference on Empirical Methods in
  Natural Language Processing and the 9th International Joint Conference on
  Natural Language Processing (EMNLP-IJCNLP)}, pages 5625--5635, Hong Kong,
  China. Association for Computational Linguistics.

\bibitem[{Toledo-Ronen et~al.(2016)Toledo-Ronen, Bar-Haim, and
  Slonim}]{toledo-ronen-etal-2016-expert}
Orith Toledo-Ronen, Roy Bar-Haim, and Noam Slonim. 2016.
\newblock \href {https://doi.org/10.18653/v1/W16-2814} {Expert stance graphs
  for computational argumentation}.
\newblock In \emph{Proceedings of the Third Workshop on Argument Mining
  ({A}rg{M}ining2016)}, pages 119--123, Berlin, Germany. Association for
  Computational Linguistics.

\bibitem[{Toledo-Ronen et~al.(2020)Toledo-Ronen, Orbach, Bilu, Spector, and
  Slonim}]{toledo-ronen-etal-2020-multilingual}
Orith Toledo-Ronen, Matan Orbach, Yonatan Bilu, Artem Spector, and Noam Slonim.
  2020.
\newblock \href {https://doi.org/10.18653/v1/2020.findings-emnlp.29}
  {Multilingual argument mining: Datasets and analysis}.
\newblock In \emph{Findings of the Association for Computational Linguistics:
  EMNLP 2020}, pages 303--317, Online. Association for Computational
  Linguistics.

\bibitem[{Toulmin(2003)}]{toulmin_uses_2003}
Stephen~E. Toulmin. 2003.
\newblock \emph{The {{Uses}} of {{Argument}}}, updated edition.
\newblock {Cambridge University Press}.

\bibitem[{Trautmann(2020)}]{trautmann-2020-aspect}
Dietrich Trautmann. 2020.
\newblock \href {https://www.aclweb.org/anthology/2020.argmining-1.5}
  {Aspect-based argument mining}.
\newblock In \emph{Proceedings of the 7th Workshop on Argument Mining}, pages
  41--52, Online. Association for Computational Linguistics.

\bibitem[{Trautmann et~al.(2020)Trautmann, Daxenberger, Stab, Schütze, and
  Gurevych}]{trautmann2020}
Dietrich Trautmann, Johannes Daxenberger, Christian Stab, Hinrich Schütze, and
  Iryna Gurevych. 2020.
\newblock \href {https://doi.org/10.1609/aaai.v34i05.6438} {Fine-grained
  argument unit recognition and classification}.
\newblock \emph{Proceedings of the AAAI Conference on Artificial Intelligence},
  34(05):9048--9056.

\bibitem[{Vreeswijk(1997)}]{vreeswijk1997abstract}
Gerard~AW Vreeswijk. 1997.
\newblock Abstract argumentation systems.
\newblock \emph{Artificial intelligence}, 90(1-2):225--279.

\bibitem[{Wachsmuth et~al.(2016)Wachsmuth, Al-Khatib, and
  Stein}]{wachsmuth-etal-2016-using}
Henning Wachsmuth, Khalid Al-Khatib, and Benno Stein. 2016.
\newblock \href {https://www.aclweb.org/anthology/C16-1158} {Using argument
  mining to assess the argumentation quality of essays}.
\newblock In \emph{Proceedings of {COLING} 2016, the 26th International
  Conference on Computational Linguistics: Technical Papers}, pages 1680--1691,
  Osaka, Japan. The COLING 2016 Organizing Committee.

\bibitem[{Wachsmuth et~al.(2017{\natexlab{a}})Wachsmuth, Naderi, Hou, Bilu,
  Prabhakaran, Thijm, Hirst, and Stein}]{wachsmuth-etal-2017-computational}
Henning Wachsmuth, Nona Naderi, Yufang Hou, Yonatan Bilu, Vinodkumar
  Prabhakaran, Tim~Alberdingk Thijm, Graeme Hirst, and Benno Stein.
  2017{\natexlab{a}}.
\newblock \href {https://www.aclweb.org/anthology/E17-1017} {Computational
  argumentation quality assessment in natural language}.
\newblock In \emph{Proceedings of the 15th Conference of the {E}uropean Chapter
  of the Association for Computational Linguistics: Volume 1, Long Papers},
  pages 176--187, Valencia, Spain. Association for Computational Linguistics.

\bibitem[{Wachsmuth et~al.(2017{\natexlab{b}})Wachsmuth, Potthast, Al-Khatib,
  Ajjour, Puschmann, Qu, Dorsch, Morari, Bevendorff, and
  Stein}]{wachsmuth:2017e}
Henning Wachsmuth, Martin Potthast, Khalid Al-Khatib, Yamen Ajjour, Jana
  Puschmann, Jiani Qu, Jonas Dorsch, Viorel Morari, Janek Bevendorff, and Benno
  Stein. 2017{\natexlab{b}}.
\newblock \href {http://aclweb.org/anthology/W17-5106} {Building an argument
  search engine for the web}.
\newblock In \emph{Proceedings of the 4th Workshop on Argument Mining}, pages
  49--59. Association for Computational Linguistics.

\bibitem[{Wachsmuth et~al.(2018)Wachsmuth, Stede, El~Baff, Al-Khatib,
  Skeppstedt, and Stein}]{wachsmuth-etal-2018-argumentation}
Henning Wachsmuth, Manfred Stede, Roxanne El~Baff, Khalid Al-Khatib, Maria
  Skeppstedt, and Benno Stein. 2018.
\newblock \href {https://www.aclweb.org/anthology/C18-1318} {Argumentation
  synthesis following rhetorical strategies}.
\newblock In \emph{Proceedings of the 27th International Conference on
  Computational Linguistics}, pages 3753--3765, Santa Fe, New Mexico, USA.
  Association for Computational Linguistics.

\bibitem[{Wachsmuth et~al.(2017{\natexlab{c}})Wachsmuth, Stein, and
  Ajjour}]{wachsmuth-etal-2017-pagerank}
Henning Wachsmuth, Benno Stein, and Yamen Ajjour. 2017{\natexlab{c}}.
\newblock \href {https://www.aclweb.org/anthology/E17-1105}
  {{``}{P}age{R}ank{''} for argument relevance}.
\newblock In \emph{Proceedings of the 15th Conference of the {E}uropean Chapter
  of the Association for Computational Linguistics: Volume 1, Long Papers},
  pages 1117--1127, Valencia, Spain. Association for Computational Linguistics.

\bibitem[{Wachsmuth et~al.(2014)Wachsmuth, Trenkmann, Stein, and
  Engels}]{wachsmuth2014modeling}
Henning Wachsmuth, Martin Trenkmann, Benno Stein, and Gregor Engels. 2014.
\newblock Modeling review argumentation for robust sentiment analysis.
\newblock In \emph{Proceedings of COLING 2014, the 25th International
  Conference on Computational Linguistics: Technical Papers}, pages 553--564.

\bibitem[{Wachsmuth and Werner(2020)}]{wachsmuth-werner-2020-intrinsic}
Henning Wachsmuth and Till Werner. 2020.
\newblock \href {https://doi.org/10.18653/v1/2020.coling-main.592} {Intrinsic
  quality assessment of arguments}.
\newblock In \emph{Proceedings of the 28th International Conference on
  Computational Linguistics}, pages 6739--6745, Barcelona, Spain (Online).
  International Committee on Computational Linguistics.

\bibitem[{Walton et~al.(2008)Walton, Reed, and Macagno}]{Reed2003-REEASI-5}
Douglas Walton, Chris Reed, and Fabrizio Macagno. 2008.
\newblock \emph{Argumentation Schemes}.
\newblock Cambridge University Press.

\bibitem[{Wang et~al.(2020)Wang, Huang, Dou, and
  Hong}]{wang-etal-2020-argumentation}
Hao Wang, Zhen Huang, Yong Dou, and Yu~Hong. 2020.
\newblock \href {https://doi.org/10.18653/v1/2020.coling-main.478}
  {Argumentation mining on essays at multi scales}.
\newblock In \emph{Proceedings of the 28th International Conference on
  Computational Linguistics}, pages 5480--5493, Barcelona, Spain (Online).
  International Committee on Computational Linguistics.

\bibitem[{Wang and Ling(2016)}]{wang-ling-2016-neural}
Lu~Wang and Wang Ling. 2016.
\newblock \href {https://doi.org/10.18653/v1/N16-1007} {Neural network-based
  abstract generation for opinions and arguments}.
\newblock In \emph{Proceedings of the 2016 Conference of the North {A}merican
  Chapter of the Association for Computational Linguistics: Human Language
  Technologies}, pages 47--57, San Diego, California. Association for
  Computational Linguistics.

\bibitem[{Wei et~al.(2016)Wei, Liu, and Li}]{wei-etal-2016-post}
Zhongyu Wei, Yang Liu, and Yi~Li. 2016.
\newblock \href {https://doi.org/10.18653/v1/P16-2032} {Is this post
  persuasive? {{R}}anking argumentative comments in online forum}.
\newblock In \emph{Proceedings of the 54th Annual Meeting of the Association
  for Computational Linguistics (Volume 2: Short Papers)}, pages 195--200,
  Berlin, Germany. Association for Computational Linguistics.

\bibitem[{Williams et~al.(2017)Williams, Nangia, and
  Bowman}]{WilliamsNB17:mnli}
Adina Williams, Nikita Nangia, and Samuel~R. Bowman. 2017.
\newblock \href {http://arxiv.org/abs/1704.05426} {A broad-coverage challenge
  corpus for sentence understanding through inference}.
\newblock \emph{CoRR}, abs/1704.05426.

\bibitem[{Yang et~al.(2019)Yang, Chen, Yang, Jurafsky, and
  Hovy}]{yang-etal-2019-lets}
Diyi Yang, Jiaao Chen, Zichao Yang, Dan Jurafsky, and Eduard Hovy. 2019.
\newblock \href {https://doi.org/10.18653/v1/N19-1364} {Let{'}s make your
  request more persuasive: Modeling persuasive strategies via semi-supervised
  neural nets on crowdfunding platforms}.
\newblock In \emph{Proceedings of the 2019 Conference of the North {A}merican
  Chapter of the Association for Computational Linguistics: Human Language
  Technologies, Volume 1 (Long and Short Papers)}, pages 3620--3630,
  Minneapolis, Minnesota. Association for Computational Linguistics.

\bibitem[{Zukerman et~al.(2000)Zukerman, McConachy, and
  George}]{zukerman-etal-2000-using}
Ingrid Zukerman, Richard McConachy, and Sarah George. 2000.
\newblock \href {https://doi.org/10.3115/1118253.1118262} {Using argumentation
  strategies in automated argument generation}.
\newblock In \emph{{INLG}{'}2000 Proceedings of the First International
  Conference on Natural Language Generation}, pages 55--62, Mitzpe Ramon,
  Israel. Association for Computational Linguistics.

\end{thebibliography}
\bibliographystyle{acl_natbib}

\end{document}